\documentclass[11pt]{article}

\usepackage[preprint]{acl}

\usepackage{times}
\usepackage{latexsym}

\usepackage[T1]{fontenc}

\usepackage[utf8]{inputenc}

\usepackage{microtype}

\usepackage{inconsolata}

\usepackage{graphicx}

\usepackage{enumitem}
\usepackage{multirow}
\usepackage{booktabs}

\title{INSEva: A Comprehensive Chinese Benchmark for Large Language Models in Insurance}

\author{
 \textbf{Shisong Chen\textsuperscript{1,2,3}},
 \textbf{Qian Zhu\textsuperscript{1}},
 \textbf{Wenyan Yang\textsuperscript{1}},
 \textbf{Chengyi Yang\textsuperscript{3}},
 \textbf{Zhong Wang\textsuperscript{1}},
 \textbf{Ping Wang\textsuperscript{1}},
\\
 \textbf{Xuan Lin\textsuperscript{1}\thanks{Corresponding author}},
 \textbf{Bo Xu\textsuperscript{4}},
 \textbf{Daqian Li\textsuperscript{1}},
 \textbf{Chao Yuan\textsuperscript{1}},
 \textbf{Licai Qi\textsuperscript{1}},
 \textbf{Wanqing Xu\textsuperscript{1}},
 \textbf{Zhenxing Sun\textsuperscript{1}},
\\
 \textbf{Xin Lu\textsuperscript{1}},
 \textbf{Shiqiang Xiong\textsuperscript{1}},
 \textbf{Chao Chen\textsuperscript{1}},
 \textbf{Haixiang Hu\textsuperscript{1}},
 \textbf{Yanghua Xiao\textsuperscript{2}}
\\
\\
 \textsuperscript{1}Ant Group,
 \textsuperscript{2}Fudan University,
 \textsuperscript{3}East China Normal University,
 \textsuperscript{4}Donghua University
\\
 \small{
   \textbf{Correspondence:} \href{mailto:daxuan.lx@antgroup.com}{daxuan.lx@antgroup.com}
 }
}

\begin{document}
\maketitle
\begin{abstract}
Insurance, as a critical component of the global financial system, demands high standards of accuracy and reliability in AI applications. While existing benchmarks evaluate AI capabilities across various domains, they often fail to capture the unique characteristics and requirements of the insurance domain. To address this gap, we present INSEva, a comprehensive Chinese benchmark specifically designed for evaluating AI systems' knowledge and capabilities in insurance. INSEva features a multi-dimensional evaluation taxonomy covering business areas, task formats, difficulty levels, and cognitive-knowledge dimension, comprising 38,704 high-quality evaluation examples sourced from authoritative materials. Our benchmark implements tailored evaluation methods for assessing both faithfulness and completeness in open-ended responses. Through extensive evaluation of 8 state-of-the-art Large Language Models (LLMs), we identify significant performance variations across different dimensions. While general LLMs demonstrate basic insurance domain competency with average scores above 80, substantial gaps remain in handling complex, real-world insurance scenarios. The benchmark will be public soon.
\end{abstract}

\section{Introduction}

\begin{figure}
    \centering
    \includegraphics[width=1.0\linewidth]{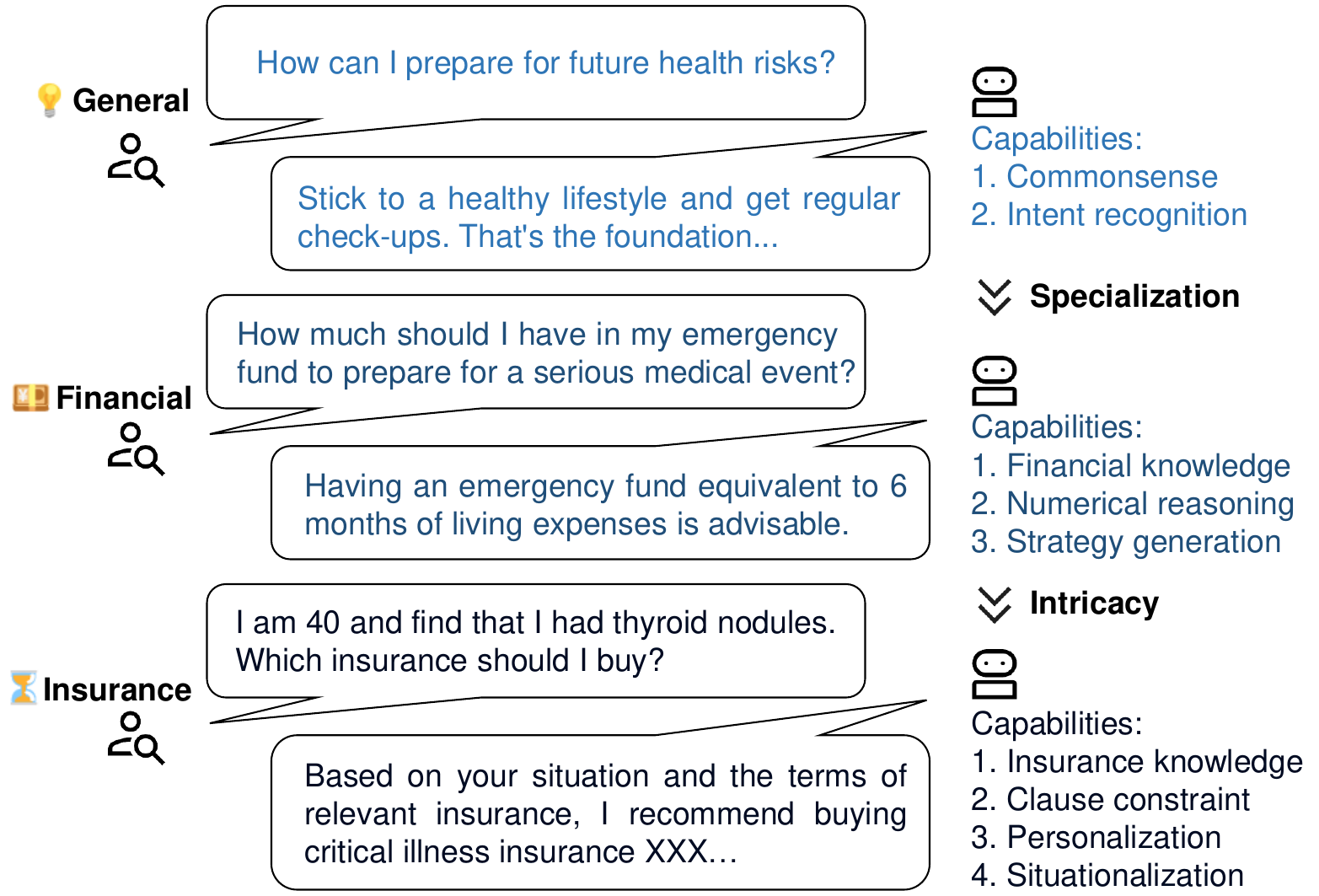}
    \caption{A comparison of question-answering examples and required model capabilities in the general, financial, and insurance domains. It shows how questions are becoming increasingly specialized and intricate in the general, financial, and insurance domains, and how this places different demands on LLM capabilities.}
    \label{fig:intro}
\end{figure}

Insurance is one of the fundamental pillars of the global financial ecosystem~\cite{pfeifer2021insurance}, performing irreplaceable infrastructure functions in economic stability worldwide. 
Recently, Artificial Intelligence (AI) has emerged as a transformative force in this field~\cite{jaiswal2023impact}. Due to the high risk of insurance decisions, AI systems are required not only to possess domain knowledge but also to ensure high accuracy and harmlessness. 
Therefore, specialized benchmarks for evaluating the domain-specific knowledge and capabilities of AI in insurance have become particularly crucial.

Many recent studies have attempted to construct benchmarks in the financial domain, which is the parent domain of insurance, yet few focus specifically on insurance. Existing financial benchmarks \cite{chen2021finqa, zhu2021tatqaquestionansweringbenchmark, chen2022convfinqa, shah2022flue, zhu2024cflue, zhang2023fineval, peng2025plutusbenchmarkinglargelanguage, xie2024finben} evaluate LLMs' capabilities within the broad financial domain.
The financial domain centers on capital circulation, value appreciation, and financial instrument transactions, emphasizing wealth growth through capital operations. Insurance, however, focuses on risk management, prioritizing risk assessment, pricing, and claims services to ensure policyholder protection and economic compensation during risk events, as shown in Figure~\ref{fig:intro}. These fundamental differences require a dedicated insurance benchmark, while currently only a few of the benchmarks are proposed for insurance.
InsQABench \cite{ding2025insqabenchbenchmarkingchineseinsurance} collects insurance commonsense question-answering data for evaluation. INS-MMBench \cite{lin2024insmmbenchcomprehensivebenchmarkevaluating} focuses on evaluating cross-modal alignment capabilities in auto insurance scenarios. These benchmarks fail to adequately cover the insurance's unique risk characteristics, complex product clauses, deep causal reasoning requirements, cross-domain knowledge integration (e.g., medical, legal), and critical elements in insurance sales processes.

To address these limitations, we present INSEva, a comprehensive Chinese benchmark specifically designed for evaluating AI systems' knowledge and capabilities in the insurance domain. Based on an in-depth study of the insurance production environment, we develop a multi-dimensional evaluation taxonomy that encompasses business areas, task formats, difficulty levels, and cognitive-knowledge domains, thereby ensuring comprehensive assessment coverage. Our data construction pipeline incorporates three key stages: (1) collection from authoritative sources, (2) systematic data augmentation to enhance diversity while maintaining domain authenticity, and (3) rigorous quality control methods involving expert validation. This methodical approach yields 38,704 high-quality evaluation examples spanning the insurance domain.
Given the high risk of insurance operations where errors can have significant financial and regulatory consequences, we implement tailored evaluation methods for different question types. For structured questions with definitive answers, we employ conventional deterministic metrics (primarily accuracy), while for open-ended responses, we deploy a specially designed LLM-based evaluation framework that systematically assesses both faithfulness to retrieved information and completeness to ground truth. 
Through these methodological innovations, INSEva not only provides a comprehensive assessment of insurance-specific capabilities but also effectively differentiates model performance across multiple dimensions. The benchmark thus offers valuable guidance for domain-specific development efforts and targeted model improvements in the domain.

Using INSEva, we evaluate 8 state-of-the-art LLMs (including closed-source, open-source, and domain-specific LLMs), revealing differentiated performance across various taxonomy dimensions and identifying numerous meaningful insights. While general LLMs demonstrate a foundational understanding of insurance concepts, as evidenced by average scores exceeding 80, a considerable gap remains between their capabilities and the expertise required to tackle complex, real-world insurance problems. Specifically, we observe a performance bottleneck in tasks requiring logical reasoning and a trade-off between faithfulness and completeness in generated content. Notably, the underperformance of the financial LLMs compared to general LLMs underscores the need for specialized benchmarks tailored to the unique demands of insurance, as existing financial benchmarks may not adequately capture the nuances of this critical domain. Furthermore, we verify the benchmark's language independence, with consistent conclusions across different languages, highlighting its minimal reliance on specific linguistic features.

To summarize, our contributions are as follows:
\begin{enumerate}
\item[-] We propose a comprehensive Chinese benchmark for the insurance domain, with a multi-dimensional evaluation taxonomy and 38,704 high-quality evaluation examples from authoritative sources.
\item[-] We design a data construction and evaluation pipeline to ensure the high-quality of data and evaluation for the insurance domain.
\item[-] We conduct systematic evaluations on several existing state-of-the-art LLMs, finding that these LLMs struggle with insurance tasks.
\end{enumerate}

\section{Related Works}

\subsection{General Benchmarks}
Many general benchmarks have been developed to assess the broad capabilities of models across diverse tasks and domains. For example, the GLUE benchmark \cite{wang2019gluemultitaskbenchmarkanalysis} comprises a collection of natural language understanding tasks. MMLU \cite{hendrycks2021measuringmassivemultitasklanguage} evaluates multi-task language understanding across a wide range of subjects, assessing both knowledge acquisition and reasoning abilities. The HELM benchmark \cite{liang2023holisticevaluationlanguagemodels} adopts a holistic approach to evaluate models across a diverse set of scenarios and metrics, emphasizing real-world relevance and ethical considerations. BIG-bench \cite{srivastava2023imitationgamequantifyingextrapolating} introduces tasks that require reasoning, common sense, and cultural knowledge. While these general benchmarks provide valuable insights into the overall performance of LLMs, they often lack the domain-specific focus and granularity required to assess performance in specialized areas such as insurance, highlighting the need for tailored evaluation resources.

\subsection{Finance \& Insurance Benchmarks}
As the parent domain of insurance, the development of financial benchmarks for financial models has accelerated in recent years, providing structured frameworks for assessing models' capabilities in the finance domain. 
FinQA \cite{chen2021finqa} advances numerical reasoning evaluation by constructing a dataset requiring multi-step calculations and inference from financial reports. 
FLUE \cite{shah2022flue} introduces comprehensive financial language understanding tasks. CFLUE \cite{zhu2024cflue} evaluates models through two main dimensions in the Chinese financial domain, including knowledge assessment and application Assessment. FinBen \cite{xie2024finben} is an extensive open-source evaluation benchmark, including 36 datasets spanning 24 financial tasks. Despite the proliferation of financial evaluation resources, they are difficult to apply directly to insurance, which presents unique terminology, regulatory frameworks, and reasoning requirements distinct from those in general finance contexts.

Compared with the financial domain, there are far fewer benchmarks specifically for insurance. InsuranceQA \cite{feng2015applyingdeeplearninganswer} is the first benchmark for insurance that contains data collected from the internet. InsQABench \cite{ding2025insqabenchbenchmarkingchineseinsurance} encompasses three specialized insurance QA tasks, including insurance commonsense knowledge, insurance structured database, and insurance unstructured document.
Ins-MMBench \cite{lin2024insmmbenchcomprehensivebenchmarkevaluating} comprises a total of 2.2k multimodal multiple-choice questions for evaluating large vision-language models. Due to the lack of a complete insurance taxonomy, these benchmarks fail to fully cover insurance tasks.

\begin{figure*}
    \centering
    \includegraphics[width=0.9\linewidth]{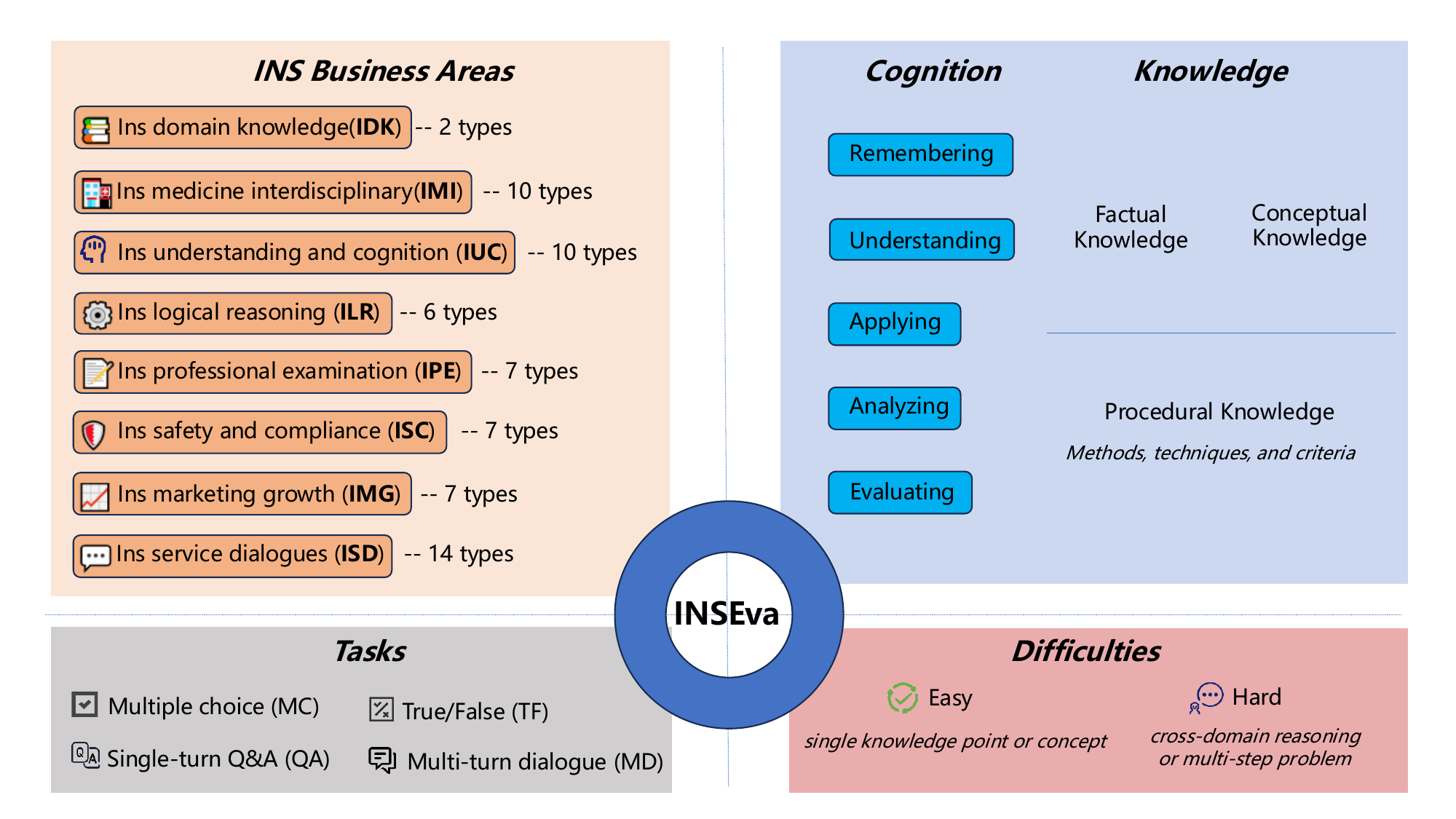}
    \caption{The taxonomy of INSEva benchmark.}
    \label{fig:tax}
\end{figure*}

\section{INSEva Bencmark}

\subsection{Overview}

INSEva is a comprehensive resource designed to evaluate the performance of models in the insurance-specific tasks. The benchmark encompasses a diverse set of samples and tasks, drawing data from authoritative Chinese sources such as professional insurance exams, regulatory standards, and real-world business data.
INSEva comprises 38,704 carefully curated examples spanning various insurance sub-domains and difficulties. Model performance is evaluated using a combination of accuracy and domain-specific metrics tailored to each task.

\subsection{Taxonomy}
\label{sec:tax}
To comprehensively assess models' capabilities in the insurance domain, we propose a multi-dimensional taxonomy for our evaluation benchmark. This taxonomy categorizes test items along four primary dimensions, including the business areas dimension, the task formats dimension, the difficulties dimension, and the cognition and knowledge dimension. The multi-dimensional taxonomy enables a fine-grained analysis of model performance across different aspects of insurance knowledge and applications.

\subsubsection{Dimension 1: Business Areas}
\label{sec:ba}
To evaluate the capabilities of LLMs in different business areas under the insurance domain, based on an in-depth study of the insurance production environment, we categorize examples according to their relevance to specific insurance business areas: (1) \textbf{Insurance domain knowledge} (IDK) examines professional knowledge reserves through 2 tasks: terminology interpretation and insurance science-related tasks; (2) \textbf{Insurance medicine interdisciplinary} (IMI) include 10 cross-domain tasks such as medical entity recognition, knowledge Q\&A, and disease prediction; (3) \textbf{Insurance understanding and cognition} (IUC) contains 10 classic NLP tasks including intent recognition and underwriting case analysis; (4) \textbf{Insurance logical reasoning} (ILR) emphasizes 6 numerical reasoning abilities, including actuarial formula derivation and numerical calculations; (5) \textbf{Insurance professional examination} (IPE) simulates 7 types of industry certification exam scenarios; (6) \textbf{Insurance safety and compliance} (ISC) establishes 7 ethical and compliance detection tasks; (7) \textbf{Insurance marketing growth} (IMG) evaluates the quality of 7 types of marketing copy generation and the effectiveness of population classification; (8) \textbf{Insurance service dialogues} (ISD) constructs 14 dialog tasks covering the entire process, including pre-investment consultation, post-investment issue handling, and professional configuration suggestions. More details are shown in the Appendix.

\subsubsection{Dimension 2: Task Formats}
\label{sec:task}
According to the task format of the example, we classify examples into four types: (1) \textbf{Multiple choice} (MC): Tasks requiring selection of the correct option(s) from a set of alternatives; (2) \textbf{True/False} (TF): Binary judgment tasks requiring verification of statements' accuracy; (3) \textbf{Single-turn Q\&A} (QA): Direct question-answer pairs requiring concise responses; (4) \textbf{Multi-turn dialogue} (MD): Extended conversations simulating real-world insurance consultations requiring contextual understanding and knowledge application across multiple exchanges.

\subsubsection{Dimension 3: Difficulties}

To assess the scalability of models across varying levels of expertise required for insurance operations, we categorize examples according to their difficulty: (1) \textbf{Easy}: Items can be answered directly without requiring deep understanding, or items only requiring the retrieval and application of a single knowledge point or concept; (2) \textbf{Hard}: Complex scenarios requiring integration and application of multiple knowledge elements, often involving cross-domain reasoning or multi-step problem-solving.

\subsubsection{Dimension 4: Cognition \& Knowledge}
\label{sec:ck}
To enable a granular assessment of models' cognitive processes and knowledge within the insurance domain, informed by established education assessment theory~\cite{krathwohl2002revision}, we categorize examples according to their alignment with specific cognitive dimensions and knowledge dimensions\footnote{It is worth noting that in the education assessment theory, there is also Creating under the cognition dimension and Metacognitive Knowledge under the knowledge dimension. Due to the seriousness and professionalism of the insurance domain, these two dimensions are not included in our taxonomy.}.

For the cognition dimension: (1) \textbf{Remembering} (Rem.): Recalling or recognizing basic insurance knowledge (e.g., facts, terminology) without deeper understanding; (2) \textbf{Understanding} (Und.): Comprehending insurance concepts through explanation, transformation, or inference to establish basic connections; (3) \textbf{Applying} (App.): Using acquired insurance knowledge or methods to solve problems in new contexts; (4) \textbf{Analyzing} (Ana.): Breaking down insurance information to clarify relationships, structures, and theories; (5) \textbf{Evaluating} (Eva.): Critically assessing the logic, structure, or effectiveness of insurance information and ideas.

For the knowledge dimension: (1) \textbf{Factual Knowledge} (FK): Basic elements of insurance (e.g., terminology, specific details); (2) \textbf{Conceptual Knowledge} (CK): Interrelationships among basic elements within insurance theory (e.g., classifications, principles); (3) \textbf{Procedural Knowledge} (PK): Methods, techniques, and criteria for determining when to use appropriate insurance procedures.

\subsection{Data Construction}

This section details our systematic approach to developing a comprehensive benchmark in the insurance domain, encompassing data collection, data augmentation, and data quality control methods.

\subsubsection{Data Collection}

Our benchmark integrates data from diverse authoritative sources to ensure comprehensiveness and domain relevance. (1) We systematically extract historical examination questions, corresponding answers, and explanatory notes from professional Chinese certification platforms, subsequently restructuring these data to conform to the format; (2) We collect specialized terminology and regulatory standards from authoritative industry repositories, from which we extract core knowledge points and reformulate them into question-answer formats using LLM; (3) We curate internal business resources containing domain-specific knowledge and scenarios, transforming them into standardized question-answer pairs using LLM. Finally, we collect about 40k examples from different sources. This multi-source approach ensures our benchmark encompasses both theoretical knowledge and practical applications within the insurance domain.

\subsubsection{Data Augmentation}

To enhance the robustness and validity of our benchmark, we implement several augmentation strategies for questions and answers on the collected data. Finally, we get nearly 80k augmented examples in the benchmark.

\paragraph{Question Augmentation} To promote linguistic diversity, we employ paraphrasing techniques to generate semantically equivalent questions with distinct syntactic expressions, thereby reducing potential model bias toward specific phrasings. Additionally, we make colloquial rewrites for standard questions, converting formal insurance terminology into colloquial expressions more representative of real-world customer interactions. This transformation enhances the benchmark's utility for customer-facing applications. 

\paragraph{Answer Augmentation} We implement option randomization to mitigate selection bias~\cite{zheng2024largelanguagemodelsrobust} for multiple-choice questions, ensuring uniform distribution of correct answers across options throughout the dataset.

\subsubsection{Data Quality Control}

To ensure the reliability and validity of our benchmark, we implement a multi-faceted quality control framework comprising rule-based, expert-based, and LLM-based validation modules.

\paragraph{Rule-Based Module} This module evaluates the quality of the samples through a series of rules. Specifically, we conduct: (1) character set validation to identify and rectify illegal characters, special symbols, or encoding anomalies, thereby ensuring Unicode compliance and eliminating corrupted elements; (2) structural validity verification for multiple-choice questions, confirming option completeness, absence of duplicates, and adherence to specified formatting requirements; and (3) content-specific quality checks, wherein samples are assessed against pre-defined criteria based on their inherent properties (e.g., questions about insurance products must explicitly reference the product name).

\paragraph{Expert-Based Module} This module ensures the faithfulness and consistency of ground truth in Q\&A and dialogue samples through rigorous domain expert examination. A team of 10 insurance industry experts, averaging five years of experience and proficient in both Chinese and English, independently reviews the data to verify that ground truth responses are fully supported by and logically consistent with the provided input context. Any inconsistent or unsupported examples are removed. Inter-annotator agreement, measured by Cohen’s Kappa, reaches 0.87, reflecting high consistency. Detailed annotation guidelines and tools are employed to ensure uniformity and reproducibility, as shown in Appendix~\ref{lab:ha}.

\paragraph{LLM-Based Module} To further enhance validation reliability and mitigate potential biases, we employ a multi-LLM voting approach. In this module, multiple distinct large language models independently evaluate each sample. Final quality determinations are then made through aggregation of these independent assessments, leveraging the diverse perspectives of multiple models to achieve a more robust and objective validation process.

\begin{table}[t]
\centering
\begin{tabular}{c|c|c|c|c} 
\hline 
\textbf{Business} & \textbf{Tasks} & \textbf{Len.} & \textbf{Num.} & \textbf{\%} \\
\hline
\hline
IDK & MC & 226 & 1336 & 3.5 \\
IMI & MC/TF & 372 & 9447 & 24.4 \\
IUC & MC & 402 & 7916 & 20.5\\
ILR & MC/TF & 603 & 3923 & 10.1\\
IPE & MC & 195 & 7932 & 20.5\\
ISC & TF & 169 & 3417 & 8.1\\
IMG & MC & 1141 & 1651 &4.3\\
ISD & QA/MD & 3230 & 3082 &8.0\\
\hline
\end{tabular}
\caption{Data statistics of different business areas in INSEva. Business areas, corresponding task types, average prompt length (in tokens), absolute counts, and relative percentage distribution are presented. The abbreviations are defined in Section~\ref{sec:tax}.}
\label{table:data_statistics_ba}
\end{table}

\begin{table}[t]
\centering
\begin{tabular}{c|c|c|c} 
\hline 
\textbf{Type} & \textbf{Tags} & \textbf{Num.} & \textbf{\%} \\
\hline
\hline
\multirow{5}{*}{Cognition} & Rem.  & 541 & 1.4 \\
 & Und.  & 8036 & 20.8 \\
 & App.  & 19339 & 50.0\\
 & Ana.  & 7038 & 18.2\\
 & Eva.  & 3750 & 9.7\\
 \hline
\multirow{3}{*}{Knowledge} & FK  & 914 & 2.4\\
 & CK  & 21047 & 54.4\\
 & PK  & 16743 & 43.3\\
\hline
\end{tabular}
\caption{Data statistics of different cognition and knowledge in INSEva. The absolute counts and relative percentage distribution are presented. The abbreviations are defined in Section~\ref{sec:tax}.}
\label{table:data_statistics_cog}
\end{table}

\subsection{Data Statistics}

Finally, we attain a total of 38,704 examples, while the average length of the prompts of these examples is 905 tokens. Table~\ref{table:data_statistics_ba} and Table~\ref{table:data_statistics_cog} show more details of different business areas and different cognition and knowledge.

\subsection{Evaluation Methods}

Due to the high risks of the insurance domain, we pay more attention to the correctness of the model's response in the insurance domain. Our evaluation framework employs different assessment approaches tailored to various question types in the benchmark. This section details the methodology for evaluating both deterministic questions with unambiguous answers and open-ended questions requiring more nuanced assessment.

For deterministic questions, such as multiple-choice and true/false questions, whether the model's response is the same as the ground-truth represents the correctness of the model's response. Specifically, we calculate \textit{Accuracy} as the primary metric, representing the proportion of questions for which the model selects the correct option from the available choices. We further stratify performance analysis across different dimensions of our taxonomy to identify specific strengths and weaknesses in model capabilities. 

For open-ended questions, including single-turn Q\&A and multi-turn dialogues, we not only require that the model's response cover the ground truth as much as possible, but also require that no hallucinations appear in the response. Therefore, we propose two metrics, faithfulness and completeness, and implement an LLM-based evaluation pipeline for each metric that approximates human expert assessment.

\begin{itemize}
    \item \textit{Faithfulness} aims to measure the consistency of a model's response with the provided context, reflecting whether the answer contains hallucinations or unsupported claims. To calculate the faithfulness, we decompose the model's response into individual statements and determine whether the provided retrieval context substantiates each statement. Specifically, we use an LLM to decompose the model's response into a maximum of 20 statements. For each statement, the LLM determines whether it is supported by the provided retrieval context, assigning a binary score of 1 if supported and 0 otherwise. The overall faithfulness score is then calculated as the proportion of supported statements.
    \item \textit{Completeness} aims to measure the extent to which a model's response covers the ground truth, reflecting whether the answer covers the valid information required. The core method involves decomposing the ground truth into statements and assessing whether each statement is present in the model's response. Specifically, we employ an LLM to decompose the ground truth into a maximum of 20 statements. The LLM then determines whether each statement is present within the model's response, assigning a binary score of 1 if present and 0 otherwise. The overall Completeness score is computed as the proportion of ground truth statements that are successfully recalled in the model's response.
\end{itemize}

\section{Experiments}

\begin{table*}[]
    \centering
    \scalebox{0.93}{\begin{tabular}{c|cccccccccccc}
    \toprule
    \multirow{2}{*}{\textbf{Models}} & \multirow{2}{*}{\textbf{IDK}} & \multirow{2}{*}{\textbf{IMI}} & \multirow{2}{*}{\textbf{IUC}} & \multirow{2}{*}{\textbf{ILR}} & \multirow{2}{*}{\textbf{IPE}} & \multirow{2}{*}{\textbf{ISC}} & \multirow{2}{*}{\textbf{IMG}} & \multicolumn{2}{c}{\textbf{ISD}} & \multirow{2}{*}{\textbf{Avg.}} \\
    &&&&&&&& \textit{Faithful.} & \textit{Complete.} &&\\
        \hline
         GPT-4o & 84.24 & 79.25 & 84.90 & 56.68 & 74.83 & 73.84 & 94.85 & 82.26 & 80.97 & 79.09  \\
         Doubao-1.5 & \textbf{90.17} & \textbf{80.88} & \textbf{85.44} & 58.75 & \textbf{87.11} & \textbf{78.35} & 94.93 & \textbf{86.86} & 82.17 & \textbf{82.74} & \\
         \hline
         Qwen2.5 & 86.95 & 77.79 & 82.90 & 65.72 & 80.23 & 80.45 & 95.17 & 78.99 & 85.36 & 81.51 \\
         Qwen-QwQ & 89.05 & 80.07 & 84.62 & 70.90 & 81.11 & 76.86 & 94.98 & 73.94 & \textbf{89.47} & 82.33\\
         Deepseek-R1 & 89.48 & 79.75 & 85.00 & \textbf{73.66} & 86.15 & 70.65 & \textbf{95.71} & 74.51 & 87.76 & 82.52 \\
         Qwen3 & 86.48 & 79.51 & 84.62 & 71.74 & 85.01 & 73.34 & 95.29 & 72.87 & 88.40 & 81.92\\
         \hline
         Fin-R1 & 80.34 & 72.48 & 76.40 & 61.90 & 67.20 & 74.56 & 87.41 & 70.70 & 81.04 & 74.67 \\
         DianJin-R1 & 87.23 & 72.59 & 82.80 & 69.36 & 82.65 & 70.19 & 85.58 & 71.21 & 87.55 & 78.80\\

    \toprule
    \end{tabular}}
    \caption{Main results for LLMs on INSEva across different business areas (abbreviations defined in Section~\ref{sec:tax}).}
    \label{tab:main_res}
\end{table*}

\begin{figure}
    \centering
    \includegraphics[width=0.9\linewidth]{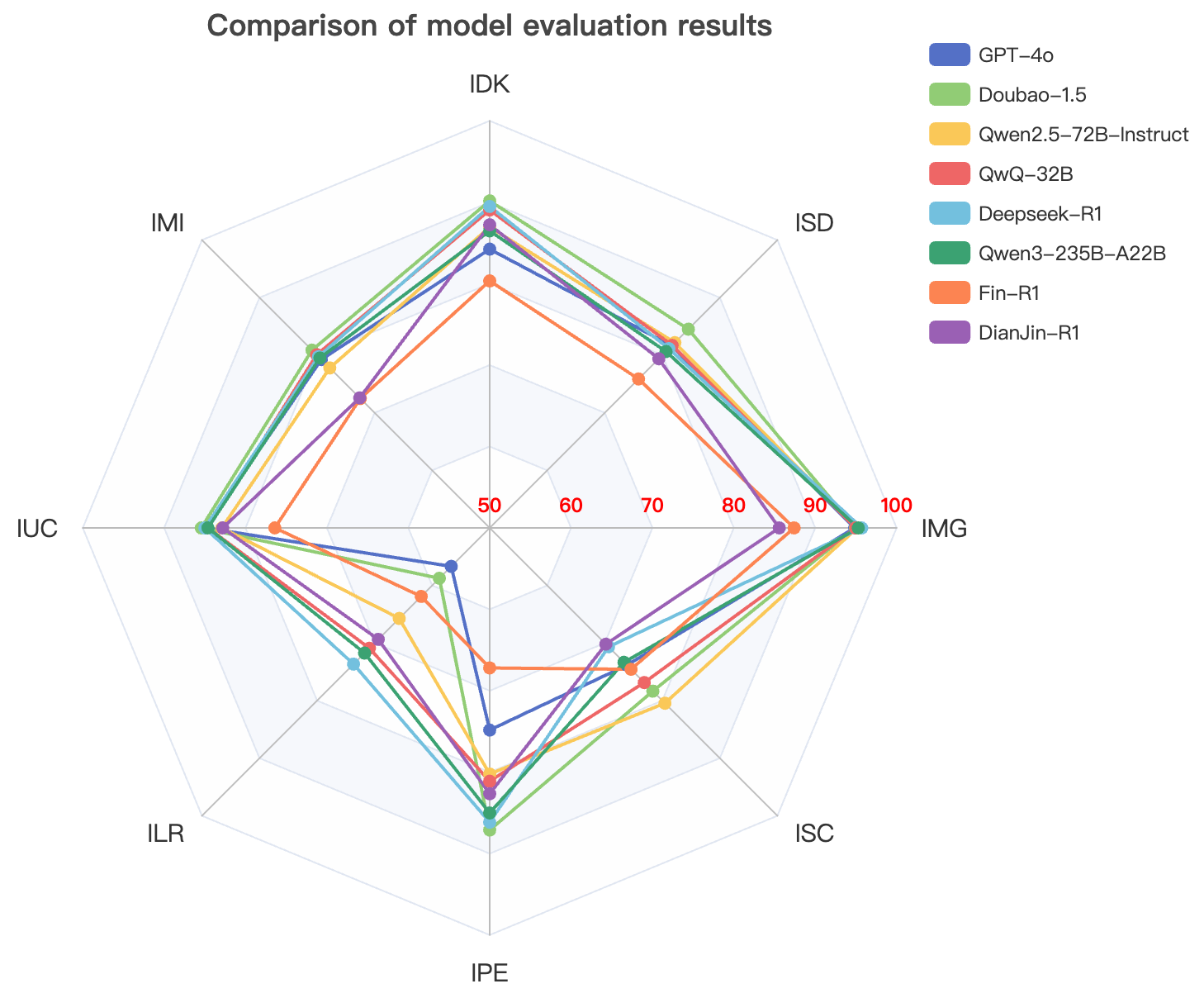}
    \caption{Performance of different LLMs across different business areas of the benchmark. The abbreviations of Business areas are defined in Section~\ref{sec:tax}.}
    \label{fig:raider}
\end{figure}

\subsection{Experimental Settings}

We select 8 state-of-the-art LLMs for evaluation, including closed-source, open-source, and domain-specific models, which are as follows: 
(1) \textbf{Closed-source:} GPT-4o~\cite{openai2024gpt4ocard}; Doubao-1.5-pro-256k\footnote{\url{https://seed.bytedance.com/en/special/doubao_1_5_pro}};
(2) \textbf{Open-source:} Qwen2.5-72B-Instruct~\cite{qwen2025qwen25technicalreport};  Qwen-QwQ-32B~\cite{qwq32b}; Deepseek-R1~\cite{guo2025deepseek}; Qwen3-235B-A22B~\cite{yang2025qwen3};
(3) \textbf{Domain-specific:} Fin-R1~\cite{liu2025finr1largelanguagemodel}; DianJin-R1~\cite{zhu2025dianjin}.

\subsection{Evaluation Results}

Based on pre-defined taxonomy and evaluation methods, we conduct a comprehensive assessment, and the results are presented in Table~\ref{tab:main_res}. A comprehensive evaluation of LLMs across different dimensions specific to the insurance domain reveals significant variations in overall performance. Doubao-1.5 demonstrates the highest aggregate performance, achieving a mean score of 82.74, while the domain-specific Fin-R1 model exhibits a comparatively lower mean score of 74.67. Notably, with the exception of the domain-specific model, the general LLMs (such as Qwen2.5, Deepseek-R1, etc.) maintain average scores above 80, indicating a relatively stable foundational capability in the insurance domain. However, a gap remains between current LLM performance and the level of expertise required to fully address complex real-world insurance problems. 

Regarding granular dimensions, all evaluated models achieve excellent results in the Insurance Marketing Growth (IMG), with an average score of more than 94 points. The reason is that although IMG is composed of insurance data, the tasks in it are similar to general tasks in NLP (such as text summarization, etc.), and require the general language understanding and generation capabilities of LLMs. However, they generally perform poorly in Insurance Logical Reasoning (ILR), with scores generally lower than other dimensions, where the highest score is only 73.66 of deepseek-R1. This uneven distribution of capabilities reflects that the current LLMs still have room for improvement in complex reasoning tasks. Additionally, the generally low scores in the Insurance Safety Compliance (ISC) also expose the limitations of the model in dealing with strong regulatory scenarios. The mediocre performance of the models in these domain-specific tasks further emphasizes the need for insurance benchmarks to fill the gap in capability evaluation in the insurance domain.

\begin{table*}[]
    \centering
    \scalebox{0.93}{\begin{tabular}{c|cccccccccc}
    \toprule
    \multirow{2}{*}{\textbf{Models}} & \multirow{2}{*}{\textbf{IDK}} & \multirow{2}{*}{\textbf{IMI}} & \multirow{2}{*}{\textbf{IUC}} & \multirow{2}{*}{\textbf{ILR}} & \multirow{2}{*}{\textbf{IPE}} & \multirow{2}{*}{\textbf{ISC}} & \multirow{2}{*}{\textbf{IMG}} & \multicolumn{2}{c}{\textbf{ISD}} & \multirow{2}{*}{\textbf{Avg.}} \\
    &&&&&&&& \textit{Faithful.} & \textit{Complete.} &\\
        \hline
        Doubao-1.5 & \textbf{86.67} & \textbf{79.38} & 82.61 & 59.16 & 74.12 & 71.15 & 90.00 & 84.45 & 80.62 & 78.68 \\
        Qwen2.5 & 84.59 & 75.85 & 81.62 & 64.73 & 68.50 & \textbf{92.82} & \textbf{94.14} & \textbf{85.51} & 71.15 & 79.88 \\
        Qwen-QwQ & \textbf{86.67} & 77.57 & \textbf{83.24} & 70.82 & \textbf{75.29} & 81.66 & 93.09 & 70.48 & \textbf{83.26} & \textbf{80.23} \\
        Qwen3 & 85.84 & 76.71 & 81.88 & \textbf{75.70} & 72.07 & 73.99 & 93.93 & 75.73 & 80.13 & 79.55 \\
        Fin-R1 & 69.58 & 67.56 & 78.15 & 71.03 & 57.05 & 89.96 & 83.69 & 62.28 & 72.23 & 72.39 \\
    \toprule
    \end{tabular}}
    \caption{Cross-Lingual evaluation results. The performance of several LLMs on a small constructed English benchmark is similar to that on Chinese.}
    \label{Tcrosslingual}
\end{table*}

Different LLMs show obvious differentiated capabilities. Doubao-1.5 performs well in all dimensions, showing strong general performance. Qwen series LLMs have relative advantages in reasoning and dialogue generation. Fin-R1 and DianJin-R1 show typical domain-specific characteristics. They have excellent performance in the financial benchmark, but their performance in insurance is even worse than that of general LLMs. It shows that even though insurance is a subdomain of finance and there are many existing financial benchmarks, it is still very important to construct an insurance benchmark.

Surprisingly, through an in-depth analysis of the faithfulness and completeness under the insurance service dialogue (ISD), we can find that different types of LLMs show significant trade-off characteristics in these two key indicators. Specifically, reasoning models represented by Deepseek-R1 perform well in completeness, but their faithfulness index is relatively low. It shows that this type of model tends to generate more comprehensive answers through reasoning and knowledge integration, which can better cover the key information points in the ground truth, but at the same time it is also more likely to generate additional content that is not explicitly mentioned in the dataset. In contrast, the non-reasoning LLMs represented by GPT-4o show relatively high faithfulness, but the completeness is relatively low, reflecting that this type of model is more inclined to generate strict answers based on known information. This trade-off relationship between faithfulness and completeness has special practical significance in high-risk financial fields such as insurance. Insurance products usually involve complex clause details and strict regulatory requirements. False or inaccurate information generated by the model may cause users to make wrong insurance decisions, leading to serious economic losses or even legal disputes. Therefore, in practical applications, we believe that faithfulness should be given priority to ensure the reliability of model output.

Main results for LLMs across different cognition and knowledge are shown in Table~\ref{tab:cog_res}. The experimental results reveal several noteworthy patterns across cognitive and knowledge dimensions. In the cognitive domain, all models demonstrate stronger performance in lower-order cognitive tasks such as Remembering (Rem.) and Understanding (Und.), with scores consistently above 85\%, while showing relative weakness in higher-order cognitive tasks, particularly in Evaluation (Eva.) where scores drop below 75\%. Notably, Deepseek-R1 exhibits exceptional performance in basic cognitive tasks (91.39\% in Remembering), but experiences a significant performance degradation in evaluation tasks (69.24\%), suggesting a common challenge in complex reasoning capabilities across all models.

\subsection{Cross-Lingual Evaluation}

We conduct several cross-lingual experiments, as shown in Table~\ref{Tcrosslingual}. Specifically, we sample approximately 10\% of the Chinese questions (around 4k) from our benchmark and translate them into English, and then test them across several models. Although the overall performance on English is slightly lower, the key conclusions remain consistent: performance is relatively low on ILR, a trade-off exists between faithfulness and completeness, and domain-specific models underperformed compared to general models. The results demonstrate that the benchmark is designed to assess the domain knowledge and reasoning abilities of LLMs in insurance, which remain consistent across languages without strong linguistic dependency.

\section{Conclusion}

We present INSEva, a comprehensive benchmark for evaluating models' capabilities in the insurance domain. Through careful design of a multi-dimensional taxonomy and rigorous data construction, we develop a benchmark of 38,704 high-quality examples.
Our evaluation of LLMs reveals several insights. While current models demonstrate basic insurance competency, they struggle with complex reasoning tasks and show a clear trade-off between faithfulness and completeness in their responses. The underperformance of financial LLMs compared to general LLMs highlights the unique challenges of the insurance domain, emphasizing the need for specialized evaluation and training approaches.
The benchmark have significant implications for both research and practical applications. For researchers, our results underscore the need to develop new architectures in high-risk domains. For practitioners, INSEva provides a valuable tool for model selection and evaluation.

\section*{Limitations}

Despite our efforts to ensure comprehensive coverage of insurance knowledge through a diverse range of examples, the current iteration of INSEva is exclusively in Chinese. This linguistic constraint may limit the direct applicability of our benchmark to models trained primarily on English or other languages, even though we have verified the language independence of the benchmark.

\bibliography{acl_latex}

\appendix

\section{Experiments}

\subsection{Main Results in Cognition \& Knowledge}

Main results for LLMs across different cognition and knowledge are shown in Table~\ref{tab:cog_res}. The experimental results reveal several noteworthy patterns across cognitive and knowledge dimensions. In the cognitive domain, all models demonstrate stronger performance in lower-order cognitive tasks such as Remembering (Rem.) and Understanding (Und.), with scores consistently above 85\%, while showing relative weakness in higher-order cognitive tasks, particularly in Evaluation (Eva.) where scores drop below 75\%. Notably, Deepseek-R1 exhibits exceptional performance in basic cognitive tasks (91.39\% in Remembering), but experiences a significant performance degradation in evaluation tasks (69.24\%), suggesting a common challenge in complex reasoning capabilities across all models.

In terms of knowledge dimensions, models generally perform better in Factual Knowledge (FK) compared to Procedural Knowledge (PK), with Deepseek-R1 achieving the highest FK score of 91.57\%. This pattern indicates that current models have a stronger grasp of basic facts and terminology than procedural operations in the insurance domain. The overall performance comparison shows Deepseek-R1 and Qwen3 leading with average scores of 84.76\% and 84.48\% respectively, significantly outperforming GPT-4o (79.72\%) and DianJin-R1 (79.44\%). This performance gap suggests that recent architectural improvements and training strategies have effectively enhanced models' capabilities across both cognitive and knowledge dimensions.

\subsection{Correlation Analysis}

To validate the effectiveness of evaluation methods for open-ended questions, we select 98 instances from different categories in the INSEva benchmark and choose 4 different models, totally 392 responses for correlation analysis. We choose the BLEU and ROUGE family of methods as our baselines, which are widely used metrics for text quality assessment. Human evaluators are asked to rate the models' responses on faithfulness and completeness. The human-rated scores are considered as the optimal evaluation method, and we conduct a correlation analysis between various common automatic evaluation methods and the proposed evaluation method. 

The correlation results are presented in Table~\ref{Tcorrlation}. It can be observed that our evaluation method exhibits the best correlation, which means our evaluation method can achieve high consistency with human evaluation.

\begin{table*}[]
    \centering
    \scalebox{0.93}{\begin{tabular}{c|ccccc|ccc|c}
    \toprule
    \multirow{2}{*}{\textbf{Models}} & \multicolumn{5}{c|}{\textbf{Cognition}} & \multicolumn{3}{c|}{\textbf{Knowledge}}& \\
     & \textbf{Rem.} & \textbf{Und.} & \textbf{App.} & \textbf{Ana.} & \textbf{Eva.} & \textbf{FK.} & \textbf{CK.} & \textbf{PK.} & \textbf{Avg.} \\
        \hline
         GPT-4o      & 80.30 & 87.74 & 80.67 & 74.95 & 71.33 & 83.57 & 82.14 & 77.13 & 79.72   \\
         Doubao-1.5  & 89.01 & 88.93 & 86.14 & 77.12 & 72.00 & 88.67 & \textbf{87.29} & 78.35 & 83.43  \\
         Deepseek-R1 & \textbf{91.39} & \textbf{89.28} & \textbf{87.30} & 81.33 & 69.24 & \textbf{91.57} & 84.44 & 83.58 & \textbf{84.76}  \\
         Qwen3       & 88.23 & 87.81 & 85.89 & \textbf{81.71} & \textbf{75.17} & 88.69 & 84.82 & \textbf{83.59} & 84.48 \\
         DianJin-R1  & 85.02 & 85.51 & 81.12 & 73.80 & 69.36 & 82.83 & 81.43 & 76.47 & 79.44\\
    \toprule
    \end{tabular}}
    \caption{Main results for LLMs across different cognition and knowledge (abbreviations defined in Section~\ref{sec:ck}).}
    \label{tab:cog_res}
\end{table*}

\begin{table}[]
    \centering
    \scalebox{1.0}{\begin{tabular}{c|ccc}
    \toprule
    \textbf{Metrics} & \textbf{Faithful.} & \textbf{Complete.} & \textbf{Avg.} \\
        \hline
         Bleu-1 & 31.47 & -18.85 & 6.31 \\
         Bleu-2 & 37.05 & -15.32  & 10.87 \\
         Bleu-4 & 44.01 & -6.07  & 18.97 \\
         Rouge-1 & 38.71 & -17.45 & 10.63   \\
         Rouge-2 & 49.53 & -1.06  & 24.24  \\
         Rouge-L & 43.86 & -7.63 & 18.12\\
        \hline
        Ours & \textbf{73.60} & \textbf{55.78} &  \textbf{64.69}\\
    \toprule
    \end{tabular}}
    \caption{Sample-level correlation of different metrics.}
    \label{Tcorrlation}
\end{table}

\section{Detailed Taxonomy}
\label{sec:dt}

Our Insurance Business Areas dimensions consists of eight main areas:

\begin{enumerate}
    \item Insurance Domain Knowledge (IDK):
    \begin{itemize}
        \item Insurance Knowledge Interpretation
        \item Insurance Science
    \end{itemize}

    \item Insurance-Medicine Interdisciplinary (IMI):
    \begin{itemize}
        \item Medical Procedure Name Extraction
        \item Medical Condition Name Extraction
        \item Disease Relevance Discrimination
        \item Insurance Hospital Name Standardization
        \item Insurance Disease \& Procedure Standardization
        \item Pet Disease Diagnosis
        \item Pet Disease Examination Q\&A
        \item Pet Medication Knowledge Q\&A
        \item Medical Prescription Disease Prediction
        \item Disease Risk Prediction in Medical Reports
    \end{itemize}

    \item Insurance Understanding and Cognition (IUC):
    \begin{itemize}
        \item Insurance Intent Understanding
        \item Insurance Product Slot Recognition
        \item Insured Object Slot Recognition
        \item Insurance Disease Slot Recognition
        \item Insurance Attribute Extraction
        \item Insurance Clause Interpretation
        \item Insurance Product Selection Analysis
        \item Insurance Liability Analysis
        \item Insurance Review Tag Recognition
        \item Insurance Product Review Classification
    \end{itemize}

    \item Insurance Logical Reasoning (ILR):
    \begin{itemize}
        \item Insurance Actuarial Science
        \item Financial Mathematics
        \item Financial Numerical Computation
        \item Insurance Prior Exemption Reasoning
        \item Insurance General Exemption Reasoning
        \item Pure Outpatient Insurance Reasoning
    \end{itemize}

    \item Insurance Professional Examination (IPE):
    \begin{itemize}
        \item Insurance Professional Qualification Exams
        \item CICE Insurance Certification
        \item Practicing Physician Qualification Exam
        \item Practicing Pharmacist Qualification Exam
        \item Practicing Veterinarian Qualification Exam
        \item Chinese Actuary Examination
        \item Insurance Salesperson Certification Exam
    \end{itemize}

    \item Insurance Safety and Compliance (ISC):
    \begin{itemize}
        \item Information Security
        \item Security Baseline
        \item Insurance Document Compliance
        \item Insurance Value System
        \item Insurance Issue Identification
        \item Insurance Fact-Checking
        \item Insurance Compliance Verification
    \end{itemize}

    \item Insurance Marketing Growth (IMG):
    \begin{itemize}
        \item Insurance Target Population Positioning
        \item Insurance Service Summary
        \item Insurance Marketing Copy Generation
        \item Insurance Product Recommendation Scripting
        \item Insurance Investor Education Scripting
        \item Insurance Population Identification \& Classification
        \begin{itemize}
            \item User Purchase Intention Classification
            \item User Cognitive Level Classification
            \item User Attitude Level
            \item Insurance Type Preference
        \end{itemize}
        \item Insurance Service Strategy Formulation
        \begin{itemize}
            \item Insurance Service Necessity Determination
            \item Insurance Service Scenario Selection
            \item Insurance Service Timing Decision
            \item Insurance Service Scenario Expression
        \end{itemize}
    \end{itemize}

    \item Insurance Service Dialogues (ISD):
    \begin{itemize}
        \item Pre-Investment Consultation
        \begin{itemize}
            \item Product Interpretation
            \item Regulatory Information Interpretation
            \item Underwriting-related
            \item Policy-related
            \item Platform-related
        \end{itemize}
        \item Post-Investment Issue Resolution
        \begin{itemize}
            \item Claims Assessment-related
            \item Claim Settlement-related
            \item Post-policy Operations
        \end{itemize}
        \item Professional Allocation Advice
        \begin{itemize}
            \item Planning and Configuration
            \item Condition-based Product Selection
            \item Product Recommendation
            \item Insurance Type Comparison
            \item Product Comparison
            \item Premium and Benefit Calculation
        \end{itemize}
    \end{itemize}
\end{enumerate}

This comprehensive taxonomy covers the full spectrum of insurance business areas, from fundamental domain knowledge to practical service applications, ensuring a thorough evaluation of AI models in the insurance domain.

\section{Details of Human Annotations}
\label{lab:ha}

The human annotations in the domain expert module and correlation analysis consist of over 10 insurance industry experts with an average of 5 years of experience, based in China. They are proficient in both Chinese and English, belong to different organizations respectively, and can utilize their industry experience along with professional reference materials to analyze and evaluate answers to the questions.

Human experts have established annotation criteria that primarily evaluate answer quality from three dimensions - factuality, accuracy, and continuity - through granular assessment of knowledge points. The evaluation scale ranges from 5 (excellent) to 0 (poor):

\begin{itemize}
    \item Factuality: Whether the response content adheres to factual truth and can be substantiated by evidence from retrieved knowledge sources.
    \item Accuracy: Whether the response demonstrates professional rigor without containing errors in knowledge, expertise, or factual information.
    \item Continuity: Whether the response maintains contextual coherence, enables smooth communication flow, and exhibits clear language expression through effective discourse transitions.
\end{itemize}

\section{Case Study}

\begin{figure*}
    \centering
    \includegraphics[width=1\linewidth]{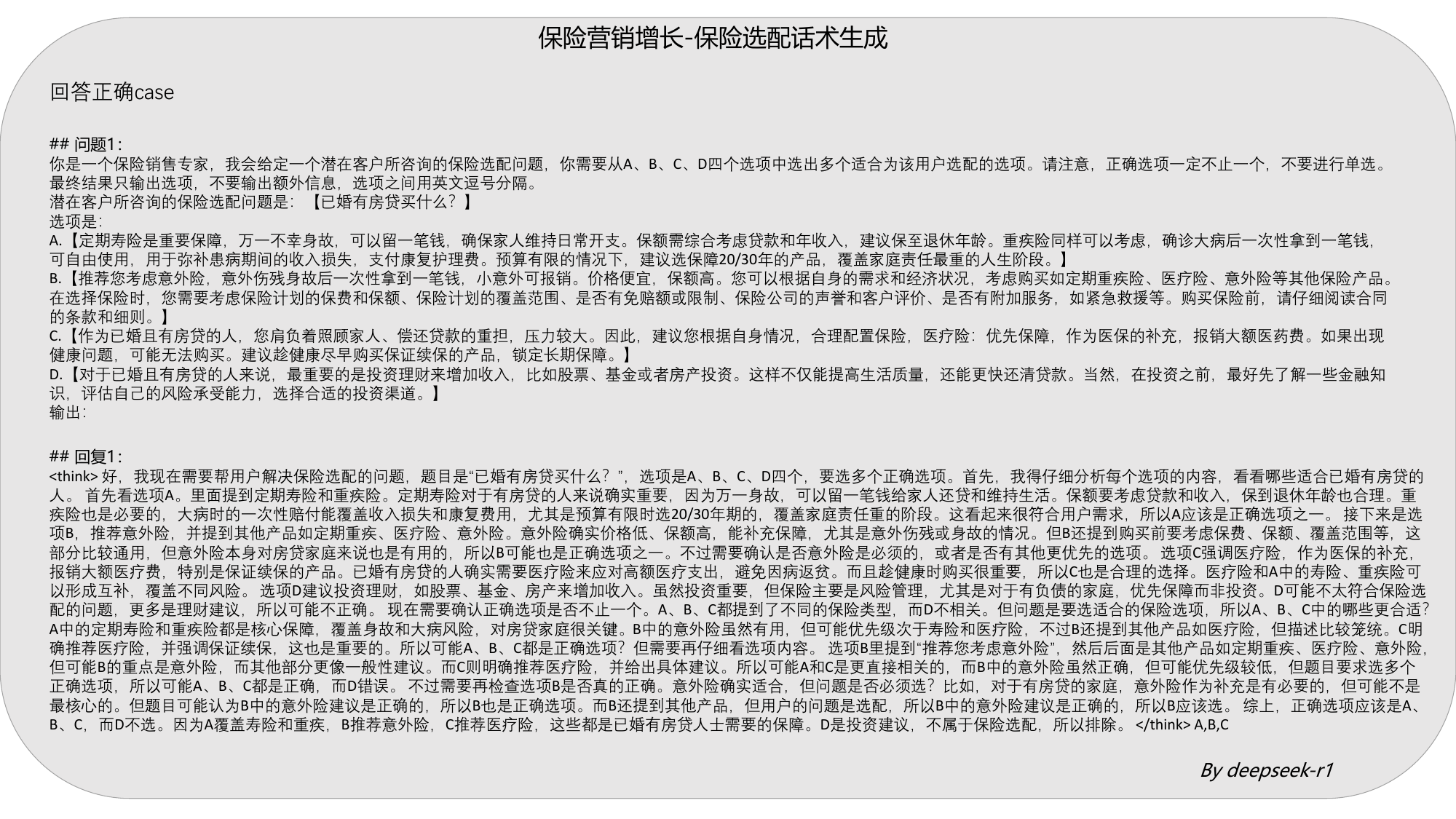}
    \caption{A case of correct answer in Chinese.}
    \label{fig:cs1}
\end{figure*}

\begin{figure*}
    \centering
    \includegraphics[width=1\linewidth]{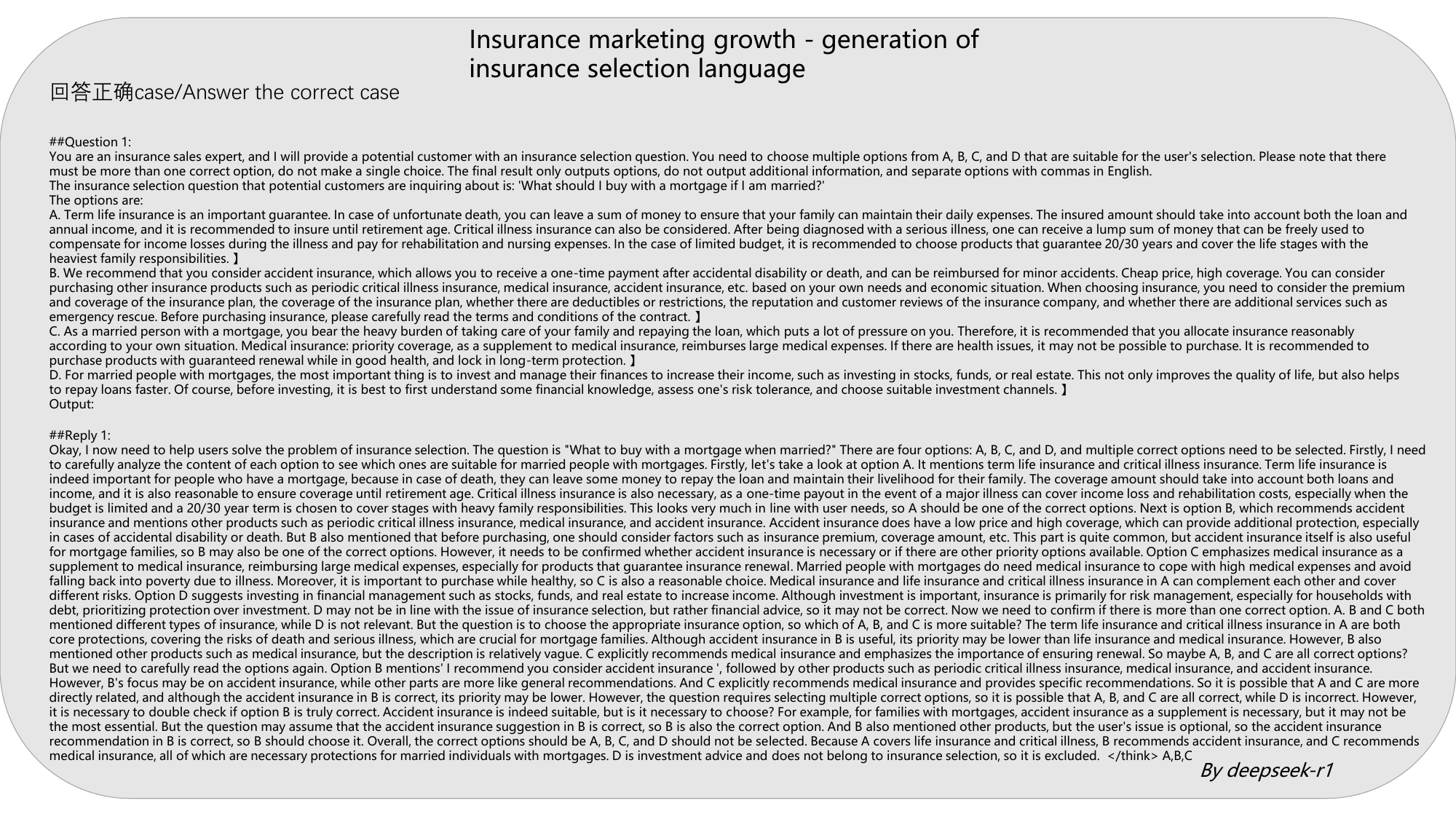}
    \caption{A case of correct answer in English.}
    \label{fig:cs2}
\end{figure*}

\begin{figure*}
    \centering
    \includegraphics[width=1\linewidth]{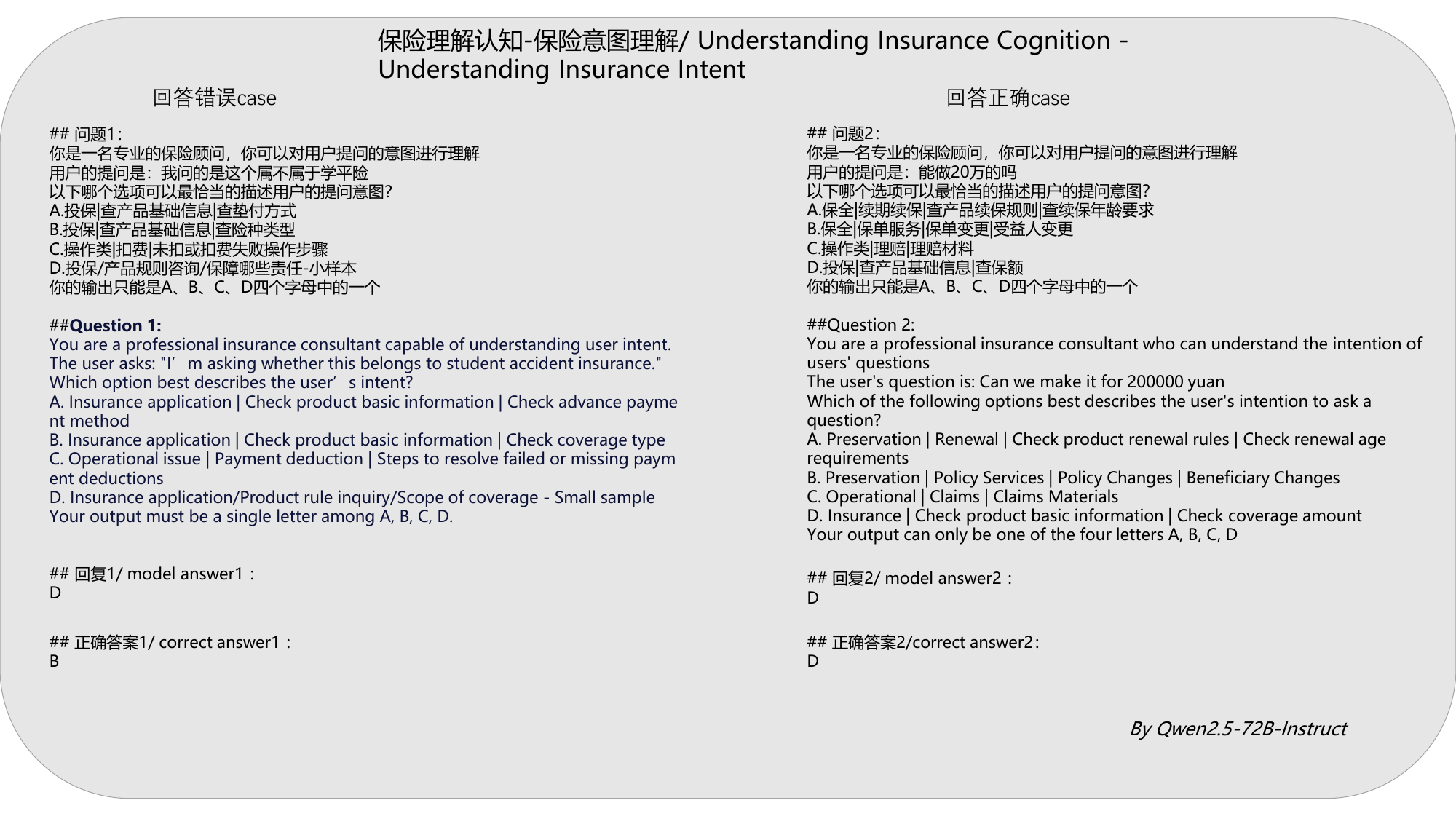}
    \caption{A case of wrong answer and correct answer.}
    \label{fig:cs3}
\end{figure*}

As illustrated in Figure~\ref{fig:cs1} and Figure~\ref{fig:cs2}, DeepSeek-R1 leverages step-by-step deliberation to correctly select the life-critical triad of term-life, critical-illness and medical coverage for a mortgaged family, evidencing its strength in multi-hop causal reasoning under complex scenarios. In contrast, Figure~\ref{fig:cs3} reveals that the non-reasoning Qwen2.5-72B-Instruct over-generalises the fine-grained intent “Does this fall under student insurance?” into a broad liability query (predicting D instead of B), yet precisely recognises a straightforward capacity question, underscoring its sensitivity to subtle semantic boundaries and the need for tailored prompting or post-calibration to mitigate misclassification.

\section{Examples}

\begin{figure*}
    \centering
    \includegraphics[width=1\linewidth]{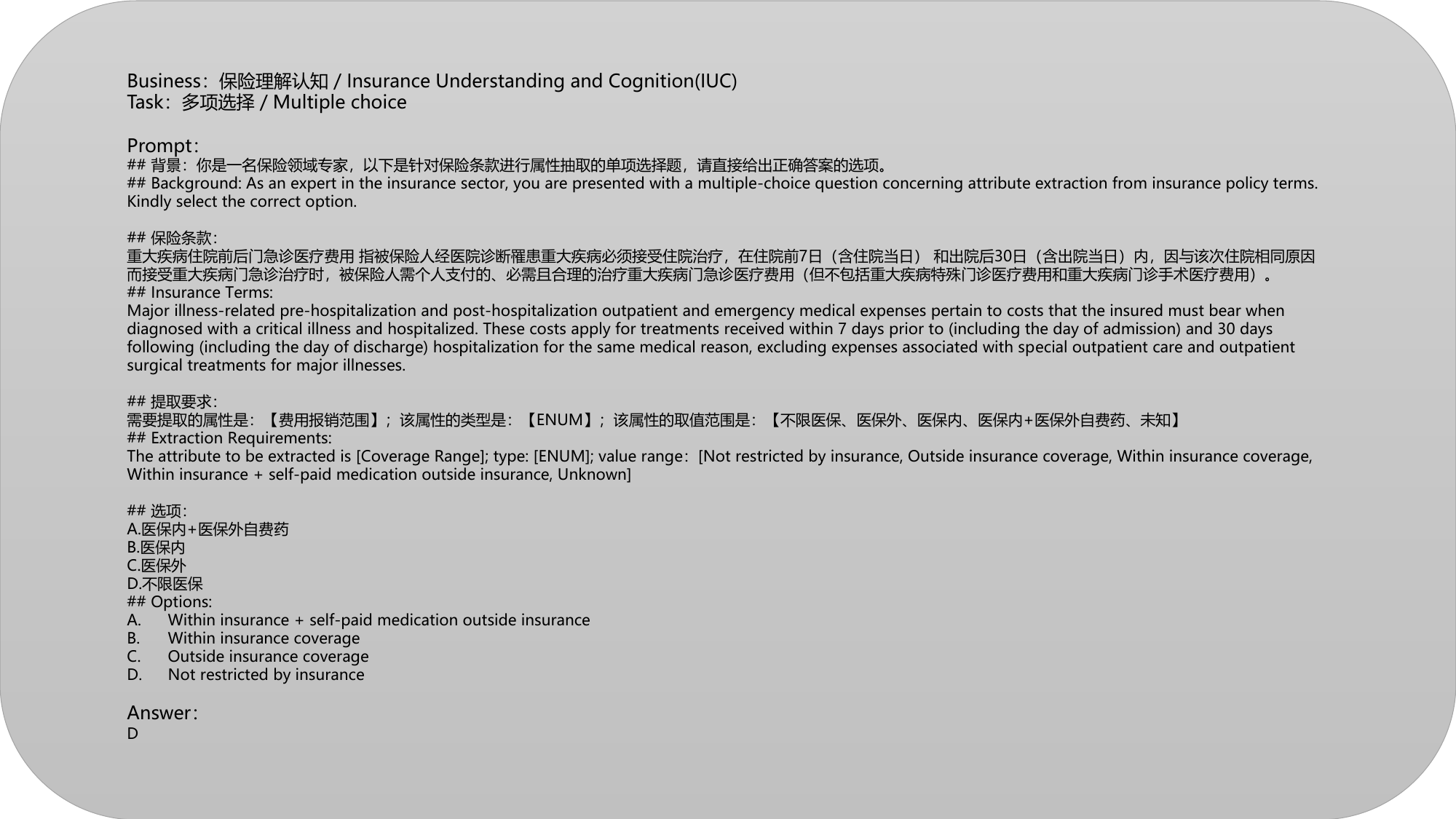}
    \caption{A multiple-choice example.}
    \label{fig:mc}
\end{figure*}

\begin{figure*}
    \centering
    \includegraphics[width=1\linewidth]{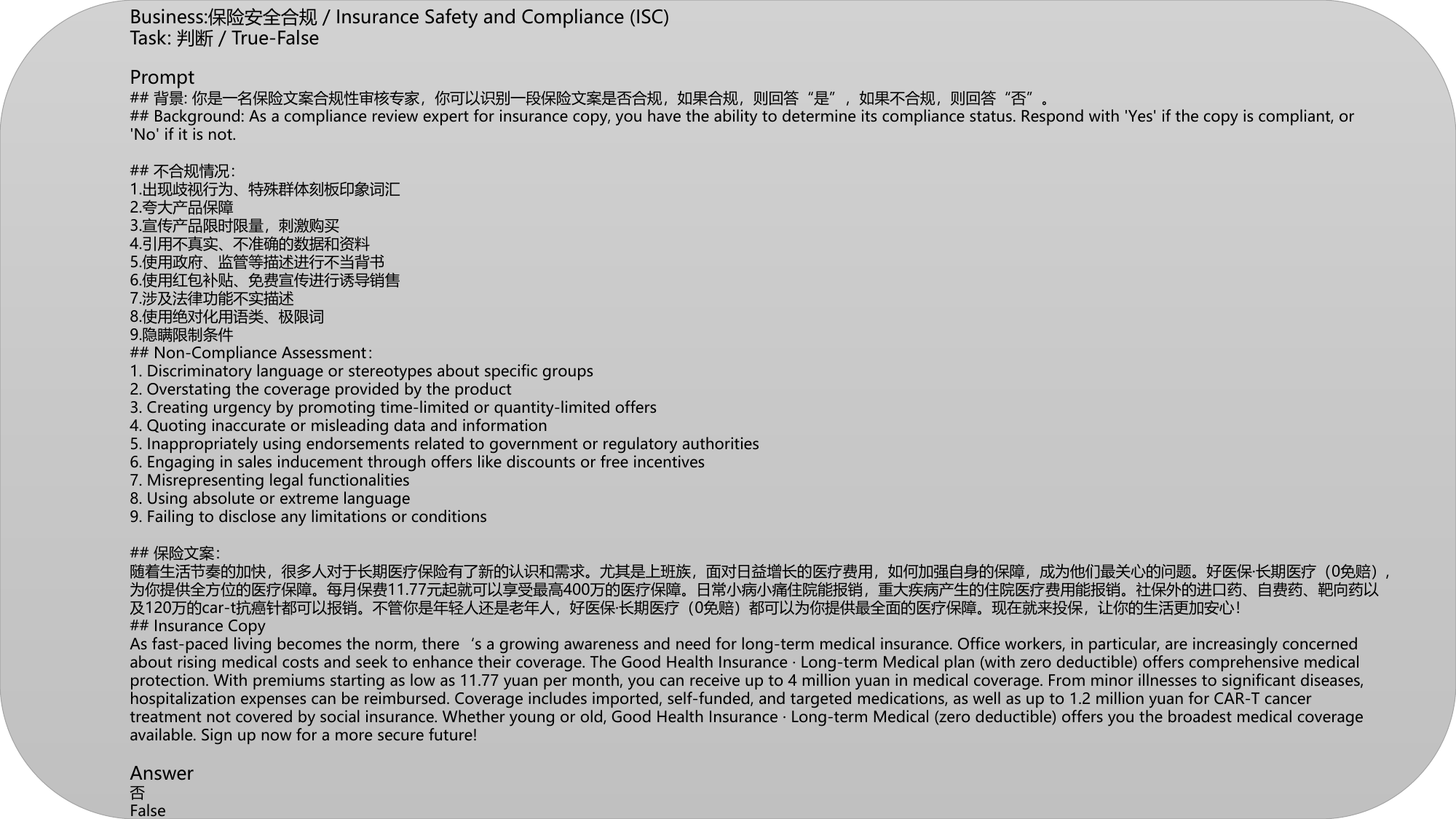}
    \caption{A True/False example.}
    \label{fig:tf}
\end{figure*}

\begin{figure*}
    \centering
    \includegraphics[width=1\linewidth]{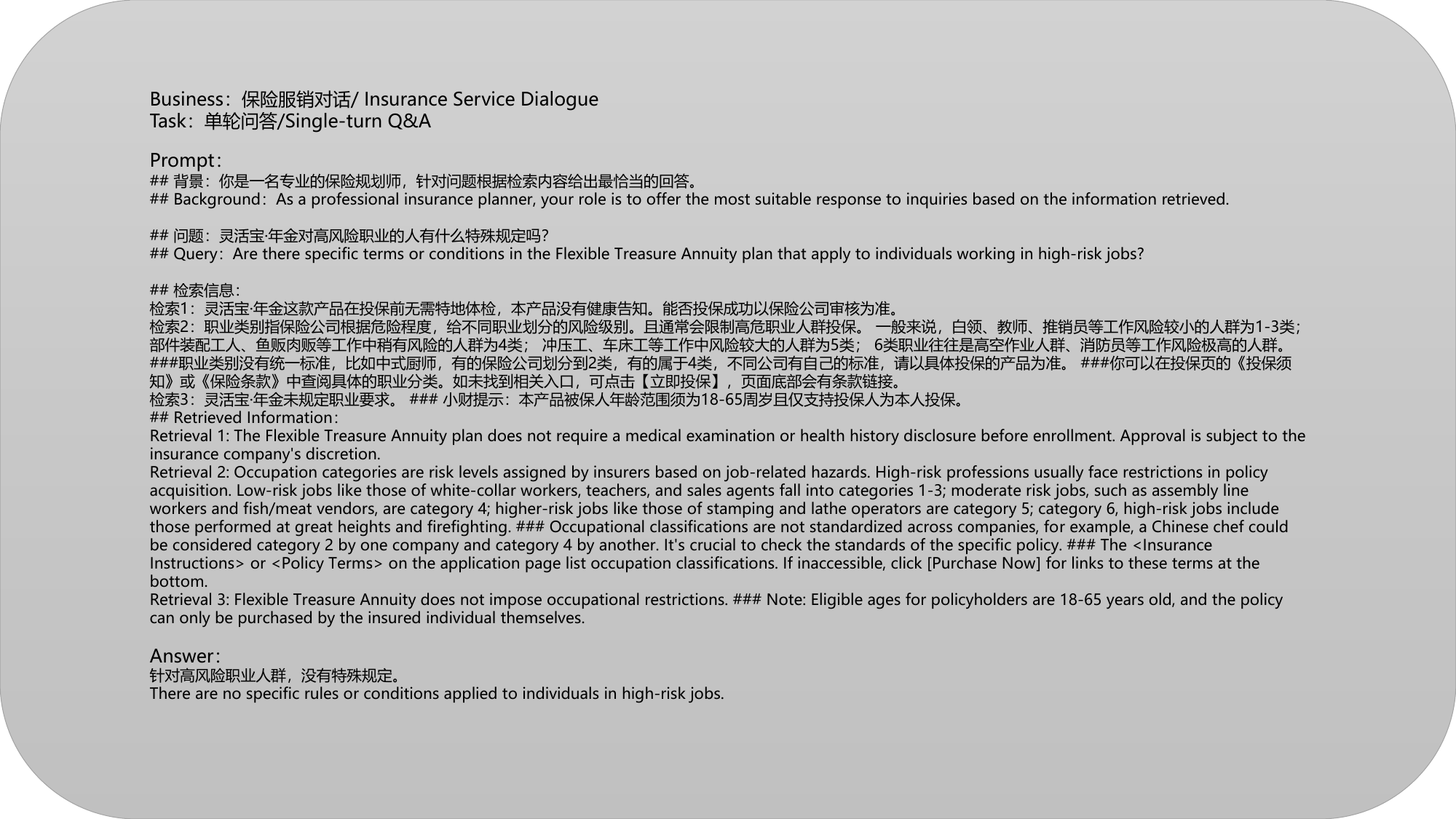}
    \caption{A single-turn Q\&A example.}
    \label{fig:qa}
\end{figure*}

\begin{figure*}
    \centering
    \includegraphics[width=1\linewidth]{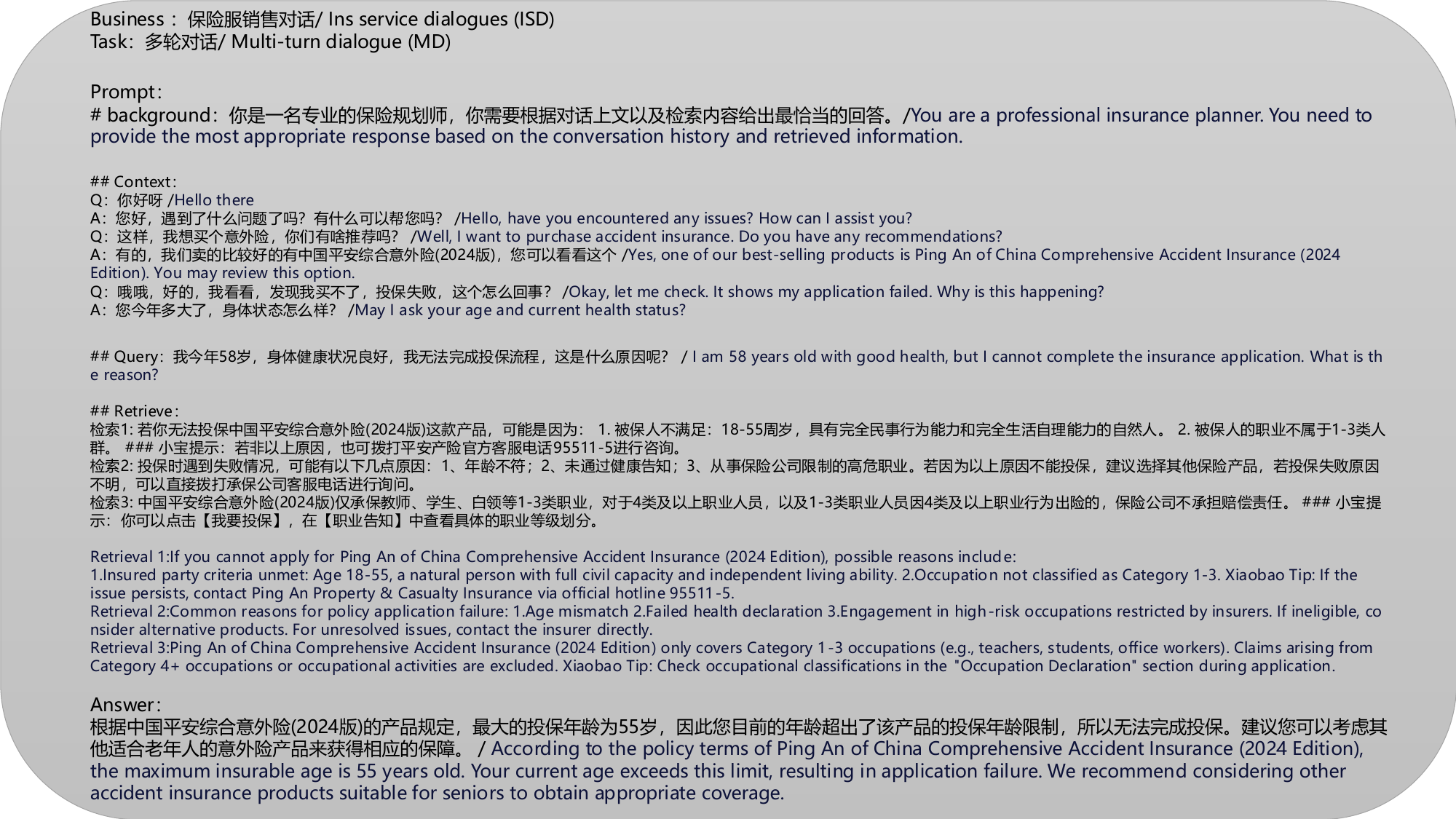}
    \caption{A multi-turn dialogue example.}
    \label{fig:md}
\end{figure*}

To illustrate the heterogeneous nature of our evaluation framework, we showcase representative examples across three distinct task formats: a multiple-choice example as demonstrated in Figure~\ref{fig:mc}, a True/False example as shown in Figure~\ref{fig:tf}, a Q\&A example as presented in Figure~\ref{fig:qa}, and a multi-turn dialogue example as presented in Figure~\ref{fig:md}.

\section{Prompts}

\subsection{Prompts For Data Construction}

\begin{figure*}
    \centering
    \includegraphics[width=1\linewidth]{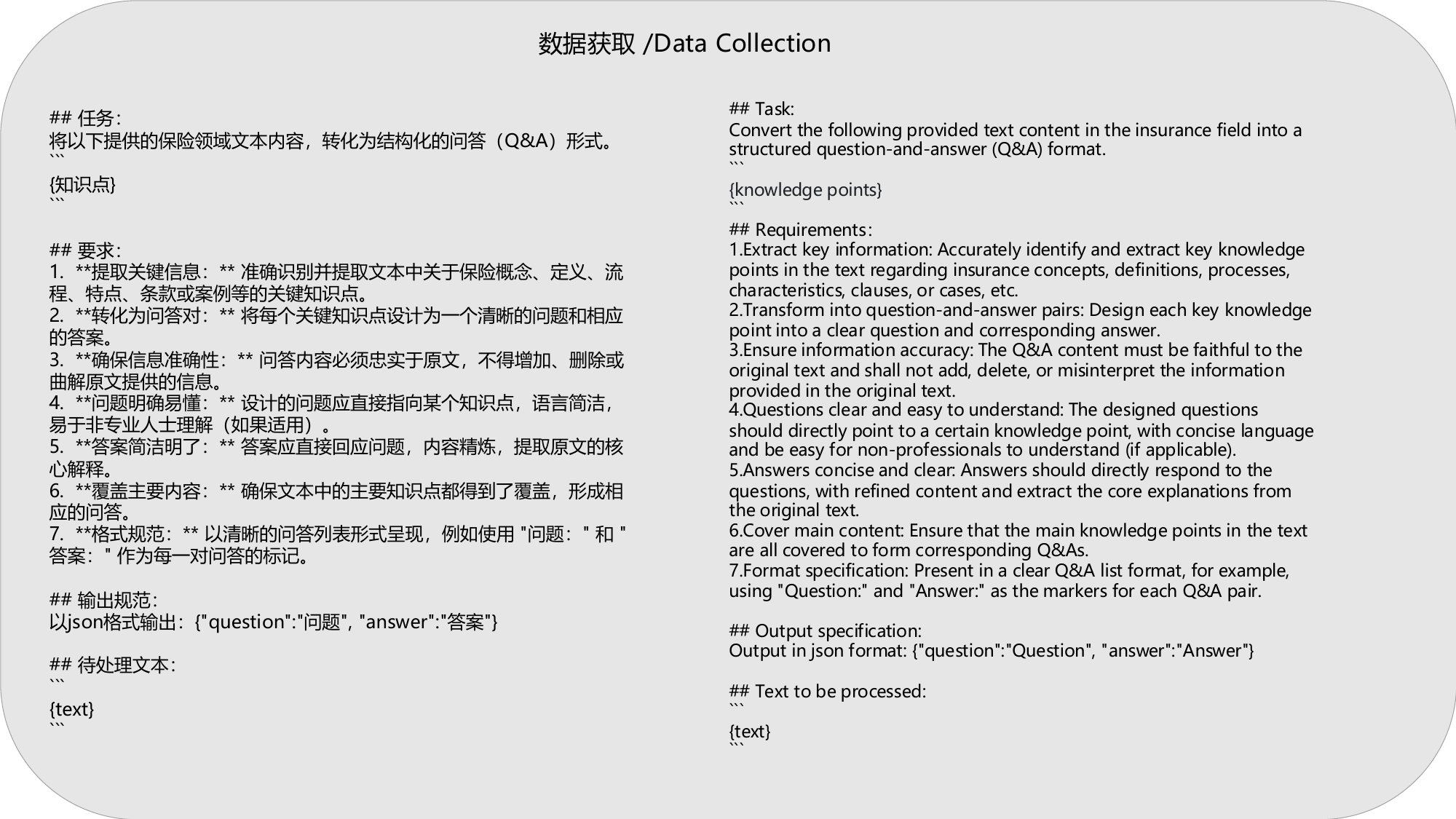}
    \caption{The prompt for converting knowledge to questions.}
    \label{fig:dg}
\end{figure*}

\begin{figure*}
    \centering
    \includegraphics[width=1\linewidth]{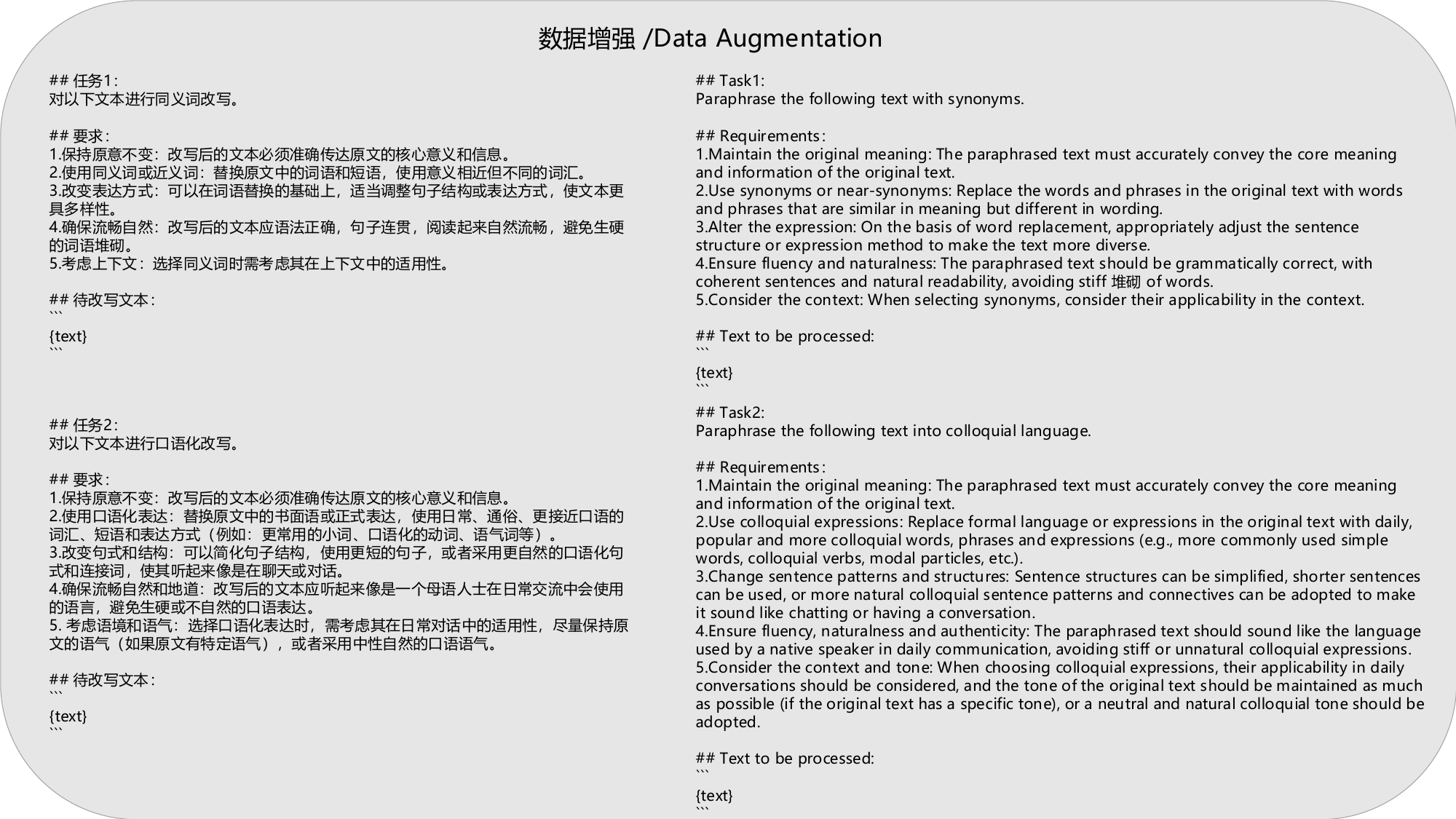}
    \caption{The prompt for data augmentation.}
    \label{fig:da}
\end{figure*}

\begin{figure*}
    \centering
    \includegraphics[width=1\linewidth]{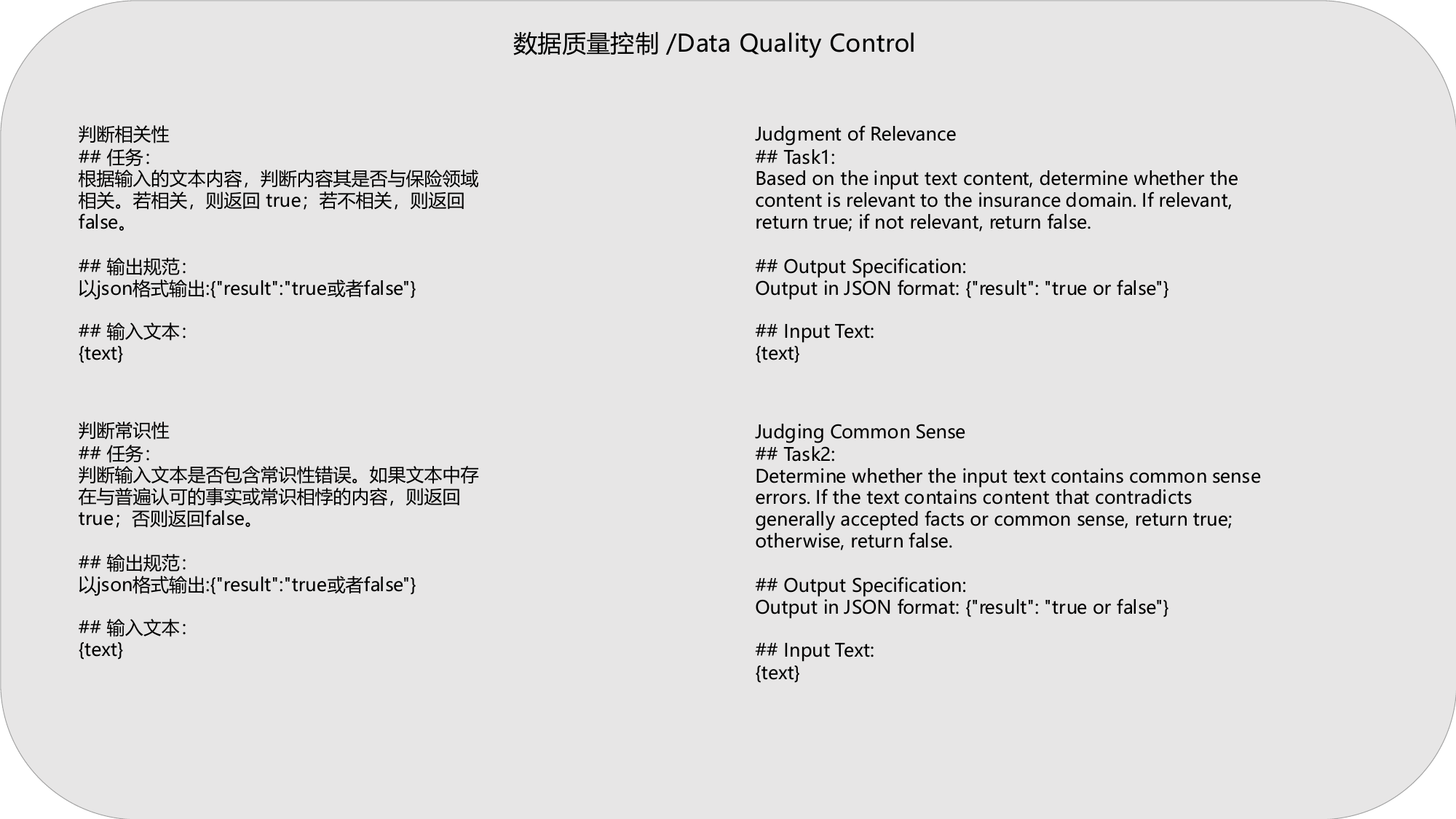}
    \caption{The prompt for data quality control.}
    \label{fig:dc}
\end{figure*}

Figure~\ref{fig:dg} shows the prompt for converting knowledge to questions.
Figure~\ref{fig:da} shows the prompt for data augmentation.
Figure~\ref{fig:dc} shows the prompt for data quality control.

\subsection{Prompts For Evaluation}

\begin{figure*}
    \centering
    \includegraphics[width=1\linewidth]{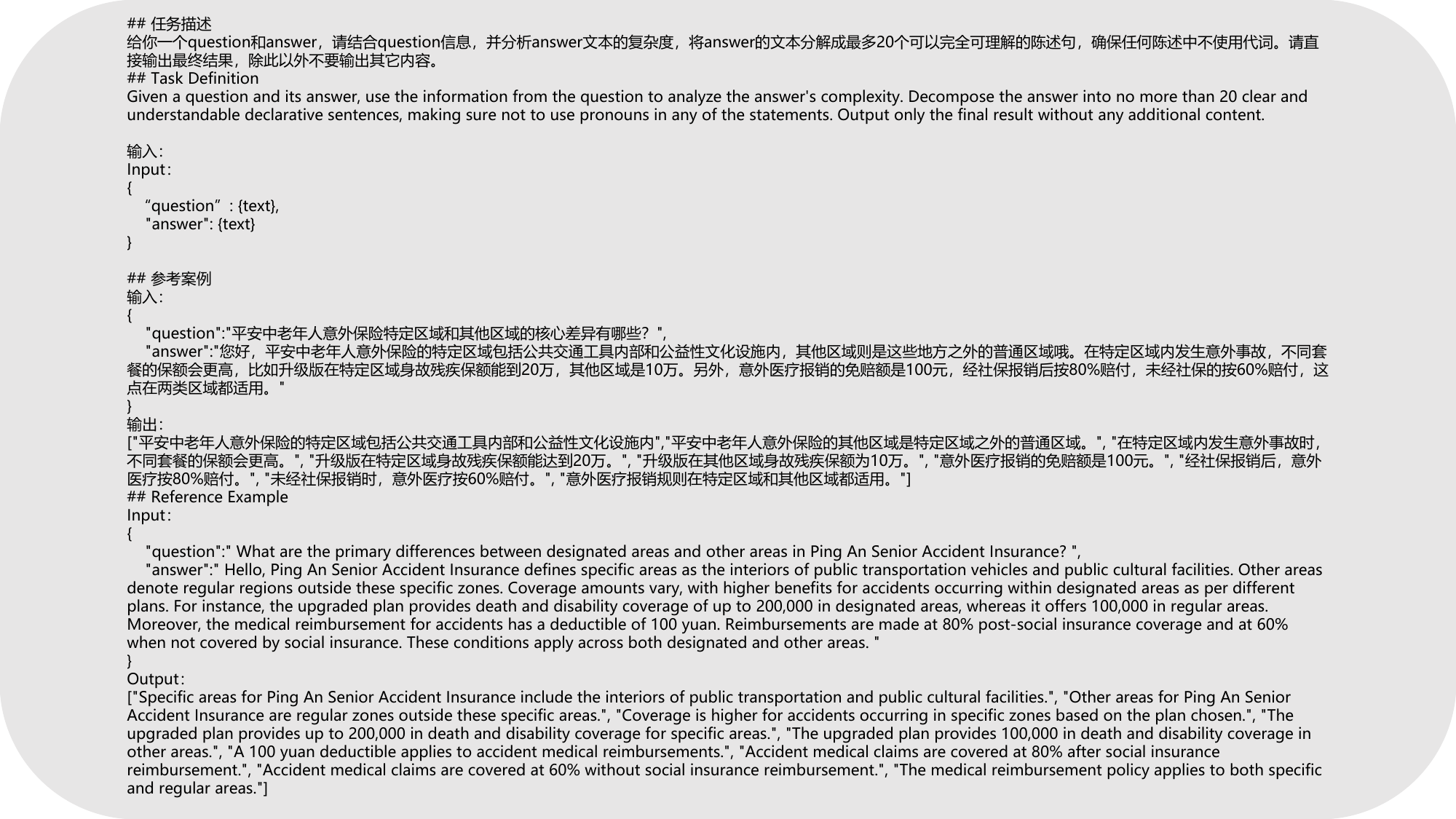}
    \caption{The prompt for answer content splitting.}
    \label{fig:sa}
\end{figure*}

\begin{figure*}
    \centering
    \includegraphics[width=1\linewidth]{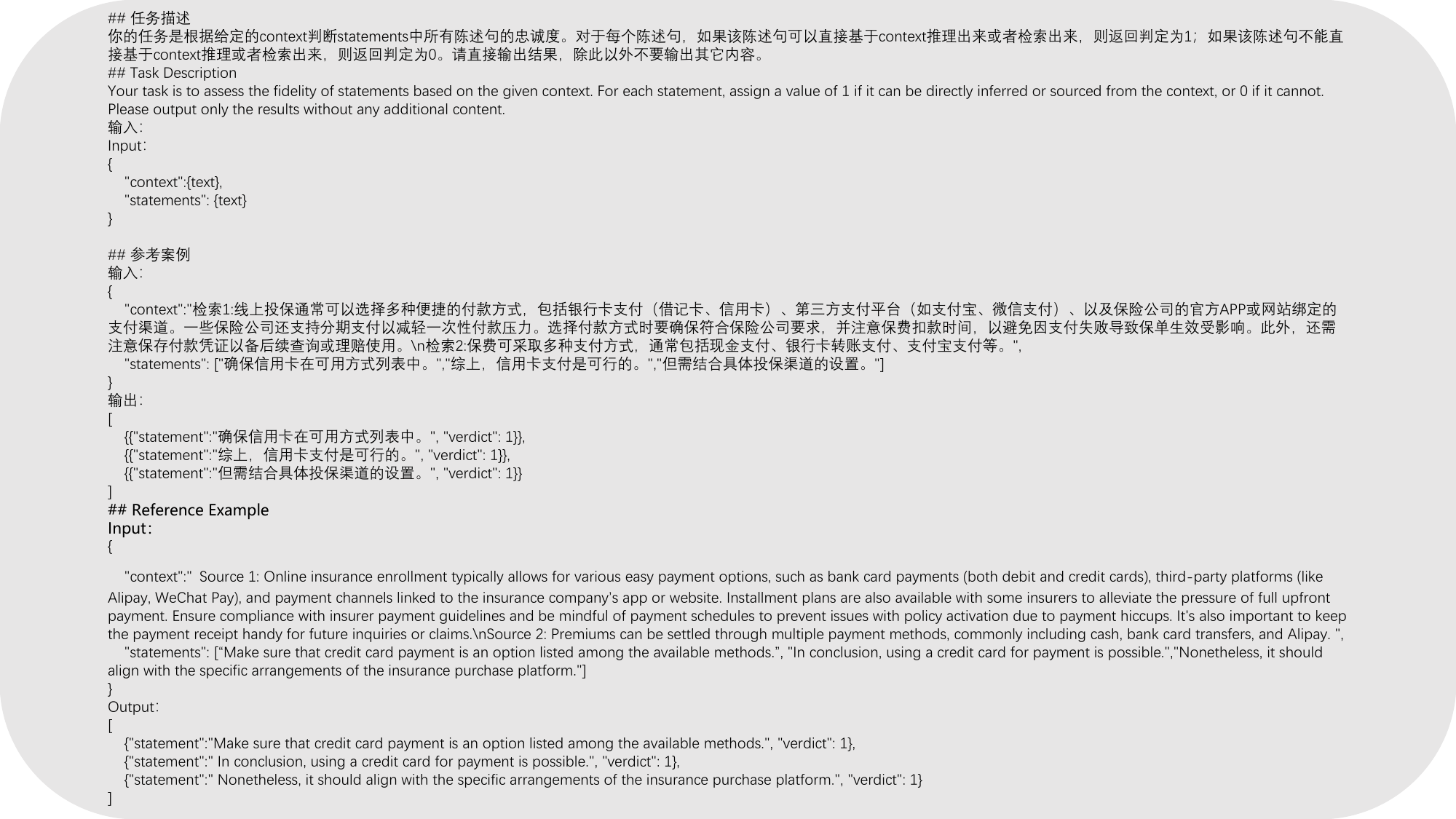}
    \caption{The prompt for answer matching.}
    \label{fig:ma}
\end{figure*}

Figure~\ref{fig:sa} and Figure~\ref{fig:ma} shows the prompt for answer content splitting and matching for calculating faithfulness.

\begin{figure*}
    \centering
    \includegraphics[width=1\linewidth]{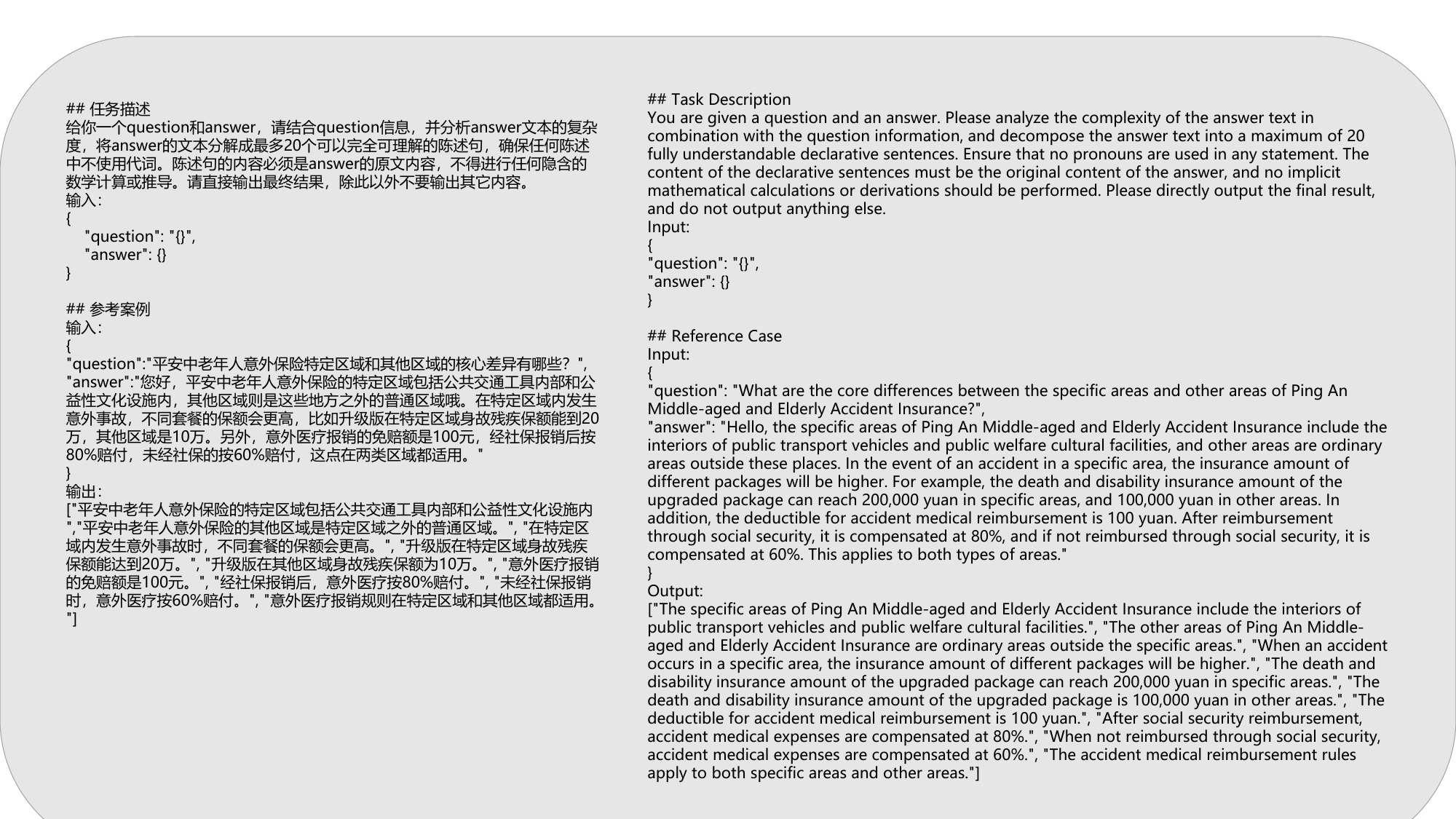}
    \caption{The prompt for ground truth content splitting}
    \label{fig:sg}
\end{figure*}

\begin{figure*}
    \centering
    \includegraphics[width=1\linewidth]{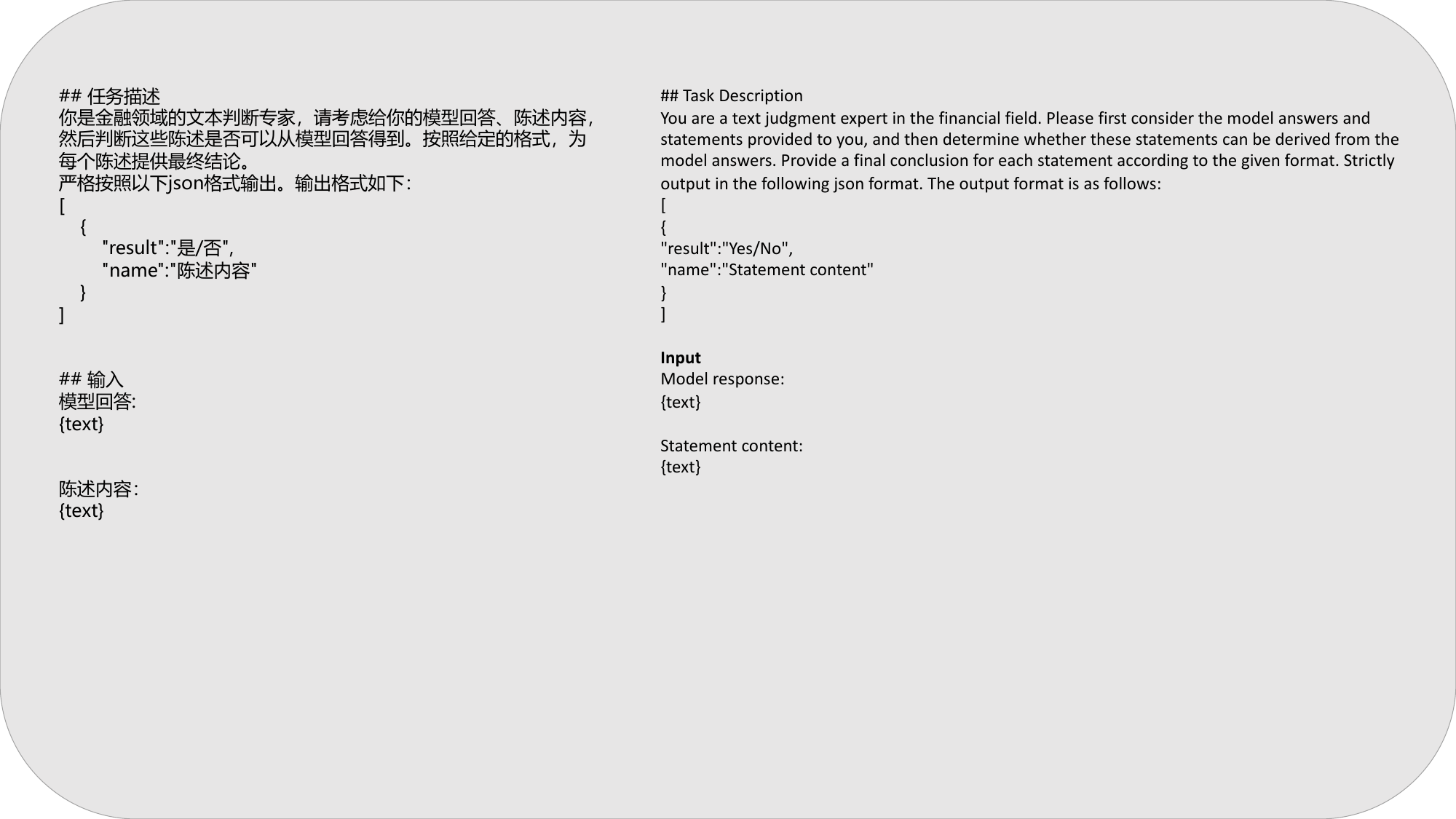}
    \caption{The prompt for ground truth matching.}
    \label{fig:mg}
\end{figure*}

Figure~\ref{fig:sg} and Figure~\ref{fig:mg} shows the prompt for ground truth content splitting and matching for calculating completeness.

\end{document}